\documentclass[12pt]{article}
\usepackage{arxiv}
\usepackage{algorithm,algorithmic}
\usepackage{amsmath,amssymb,amsfonts}

\usepackage{amsmath,amsfonts,bm}

\def\eqref#1{equation~\ref{#1}}

\def\1{\bm{1}}

\def\rvu{{\mathbf{i}}}

\def\rvs{{\mathbf{s}}}

\def\rvu{{\mathbf{u}}}

\def\rvx{{\mathbf{x}}}

\def\rvz{{\mathbf{z}}}

\def\vu{{\bm{u}}}

\def\vx{{\bm{x}}}

\def\vz{{\bm{z}}}

\def\mF{{\bm{F}}}

\def\mH{{\bm{H}}}
\def\mI{{\bm{I}}}

\def\mK{{\bm{K}}}

\def\mM{{\bm{M}}}

\def\mP{{\bm{P}}}
\def\mQ{{\bm{Q}}}
\def\mR{{\bm{R}}}

\DeclareMathAlphabet{\mathsfit}{\encodingdefault}{\sfdefault}{m}{sl}
\SetMathAlphabet{\mathsfit}{bold}{\encodingdefault}{\sfdefault}{bx}{n}

\newcommand{\E}{\mathbb{E}}

\usepackage{todonotes}
\usepackage{float}
\usepackage{multicol,multirow}
\usepackage{secdot}
\usepackage{amsmath,array,ragged2e}
\newcolumntype{C}{>{\Centering\arraybackslash}m{0.14\linewidth}}
\usepackage{booktabs}
\usepackage{adjustbox}
\usepackage{subcaption}
\usepackage{graphicx}
\usepackage{hyperref}
\graphicspath{{./pics/}}
\usepackage{lineno}

\def\eqref#1{(\ref{#1})}

\begin{document}

\title{Multiparticle Kalman filter for object localization in symmetric environments}

\author{Roman Korkin \\ 
Novosibirsk Technology Center \\
Schlumberger, Russia\\
\url{korkin.rv@phystech.edu}
\And 
Ivan Oseledets \\
Skoltech, Moscow, Russia \\
AIRI, Moscow, Russia \\
\url{i.oseledets@skoltech.ru}
\And Aleksandr Katrutsa\thanks{Corresponding author}\\
Skoltech, Moscow, Russia \\
AIRI, Moscow, Russia \\
\url{aleksandr.katrutsa@phystech.edu}
}


\maketitle

\begin{abstract}
This study considers the object localization problem and proposes a novel multiparticle Kalman filter to solve it in complex and symmetric environments.
Two well-known classes of filtering algorithms to solve the localization problem are Kalman filter-based methods and particle filter-based methods.
We consider these classes, demonstrate their complementary properties, and propose a novel filtering algorithm that takes the best from two classes.
We evaluate the multiparticle Kalman filter in symmetric and noisy environments.
Such environments are especially challenging for both classes of classical methods.
We compare the proposed approach with the  particle filter since only this method is feasible if the initial state is unknown.
In the considered challenging environments, our method outperforms the particle filter in terms of both localization error and runtime.


\end{abstract}




\section{Introduction}
Object localization problem is the problem of estimating an object’s state in an environment from sensor data and a map of the environment.
The object state may contain coordinates, velocities, internal states, and other quantities describing the internal object features.
%
The problem appears in many domains, firstly in navigation~\cite{Kalman, Lynch, Chot}, but also in image processing~\cite{pf_img_proc}, finance~\cite{Racicot} and fatigue predictions~\cite{fatigue}.

In particular, mobile cleaning robots have to solve in-door or open space localization problem~\cite{Xiao, Thrun, Huang} to perform their basic functions such as vacuum cleaning. 
Even the simple vacuum cleaning robot Roomba~\cite{Roomba} uses multiple types of sensors to localize itself in the working space.
The same localization problem appears in the development of self-driving cars~\cite{Levinson}, which use cameras, radars, LIDARs~\cite{hecht2018lidar,faizullin2022open}, 2D laser scanners, a global positioning system (GPS), and an inertial measurement system~\cite{Samyeul}. 


Despite significant differences between applications listed above, they often can be treated within a similar framework, which contains prediction and update stages.
At the prediction stage, the object state model predicts the state in the next time moment or is being initialized. 
Then, the measurements are performed with sensors. 
Based on the measurement results, the predicted object state is recomputed in the update step.


One of the classical approaches to solving the localization problem is the Kalman filter~\cite{Kalman}.
Kalman filter is a very fast and memory-efficient approach, though it has many limitations. 
For example, this method assumes  linearity of motion and measurement equations, Gaussian distribution of motion and measurement noise, and approximate knowledge of the initial object state. 
In the real-world scenario, these assumptions might not hold. 
Therefore, some modifications of the Kalman filter are used in practice, e.g. extended~\cite{julier1997new}, unscented~\cite{Julier}, and invariant extended~\cite{ieKF} Kalman filters.
They address linearity constraints by linearizing the equations around the current state estimate.
The effects from the non-Gaussian noise are treated with the entropy optimization technique in~\cite{Zhang}. 
Also, the ensembled Kalman filter~\cite{enKF} is suggested for problems with high-dimensional state vectors.
However, unlike the regular approach, the variations of the Kalman filter, generally, are not optimal estimators and even may diverge~\cite{EKF_divergence}. 

An alternative to the Kalman filter is the particle filter~\cite{PF1996, del1997nonlinear, Kunsch}, which successfully treats non-linear motion and measurement equations and non-Gaussian motion and measurement noise. 
It also works well with significant uncertainties of the initial conditions. 
However, the single iteration of the particle filter is more costly compared to the one iteration of the Kalman filter. 
Moreover, the number of particles has to be exponentially increased with the dimension of state~\cite{kf_overview}, and thus particle filter is not appropriate for solving high-dimensional problems.
In addition, the method is purely stochastic and may require too many particles for convergence, especially in symmetric environments.

The main features of the considered environments are the positions of beacons and obstacles in the space.  
The localization problem in such symmetric environments becomes especially challenging if the initial states of an object are not known. 
We show that multiple symmetrically located beacons lead to poor performance of the particle filter method. 
The symmetry in the obstacles and beacons' positions leads to the instability of the prediction results for both Kalman and particle filters.
This phenomenon is observed if the noise in measurement results leads to the ambiguity of the location among symmetrical subparts.
Such symmetric environments model real-world settings, e.g. in-door navigation in standardized buildings and in symmetrically arranged city blocks.


To address the excessive number of particles in symmetric environments, we suggest a natural combination of the particle and Kalman filters.
This combination is further referred to as the Multiparticle Kalman filter (MKF) and is based on the following ideas.
Each particle is processed with the Kalman filter equations and then, particle weights are updated based on the particle filter approach.
On the one hand, such a combination has a higher per-iteration complexity compared to the particle filter.
On the other hand, using Kalman equations in the prediction of every particle state can lead to faster convergence and higher robustness, which is shown in Section~\ref{sec::experiments}.

The contributions of this study are the following.
\begin{itemize}
    \item We have illustrated the performance degradation of the particle filter in symmetric environments.
    \item We have developed an accurate and robust localization algorithm based on the combination of the Kalman and particle filters.
    \item We have shown that our algorithm outperforms the particle filter in terms of both localization error and runtime.
\end{itemize}

 

\paragraph{Related works.}


There are multiple combinations of the particle filter (PF) with other methods~\cite{Rao, Reconcillation, particle_swarm, particle_genetic, Tutorial, SHARIATI201932}.
Rao-Blackwellised PF~\cite{Rao} splits state vectors into two parts. 
The first part is processed with the Kalman filter, and the second part is processed with the PF. 
This approach is applicable for high dimensional problems, where the standard particle filter may fail.
The Box PF method~\cite{Interval} describes the state vector distribution as a sum of uniform probability density functions. 
This method decreases the number of particles resulting in the same accuracy as PF.
In study~\cite{Reconcillation} the measurement test criterion and data reconciliation are proposed to derive reliable initial states under sufficient information about measurements.

Other studies are devoted to a combination of the particle filter with nature-inspired optimization methods like particle swarm method~\cite{particle_swarm} or genetic algorithm~\cite{particle_genetic}.
Such combinations incorporate elements of these optimization methods into the particle filter, e.g. stochastic resampling is replaced by the crossover and mutation operations. 
Although these combinations provide better localization accuracy, they are computationally expensive.

There are also various combinations of Kalman and particle filters. 
In~\cite{Tutorial} extended Kalman particle filter is presented.
This filtering algorithm reduces uncertainty in every particle motion due to the additional Kalman-type update for every particle. 
However, 
it requires a large number of particles and converges slowly if the state noise model is incorrect.
Another combination is proposed in~\cite{SHARIATI201932} and it is used diagonal process covariance matrices fitted from experimental data.
Therefore, it requires more data and can not treat an arbitrary localization problem.

\section{Problem statement}
\label{sec::problem_statement}
Let $\rvx_t \in \mathbb{R}^d$ be a state at time $t$ described by $d$-dimensional vectors and $\rvz_t \in \mathbb{R}^k$ be the $k$-dimensional measurements results. 
For example, the object location on the plane $x, y$ and its heading $\phi$ can be considered as $3D$ state, while the results of distance measurements to the $k$ nearest beacons can be considered as $kD$ measurement vector.

\begin{figure}[!h]
    \hspace{-0cm}
    \hspace{-2cm}
    \includegraphics[width=15cm, trim={0cm 0cm 0cm 0cm}, clip]
    {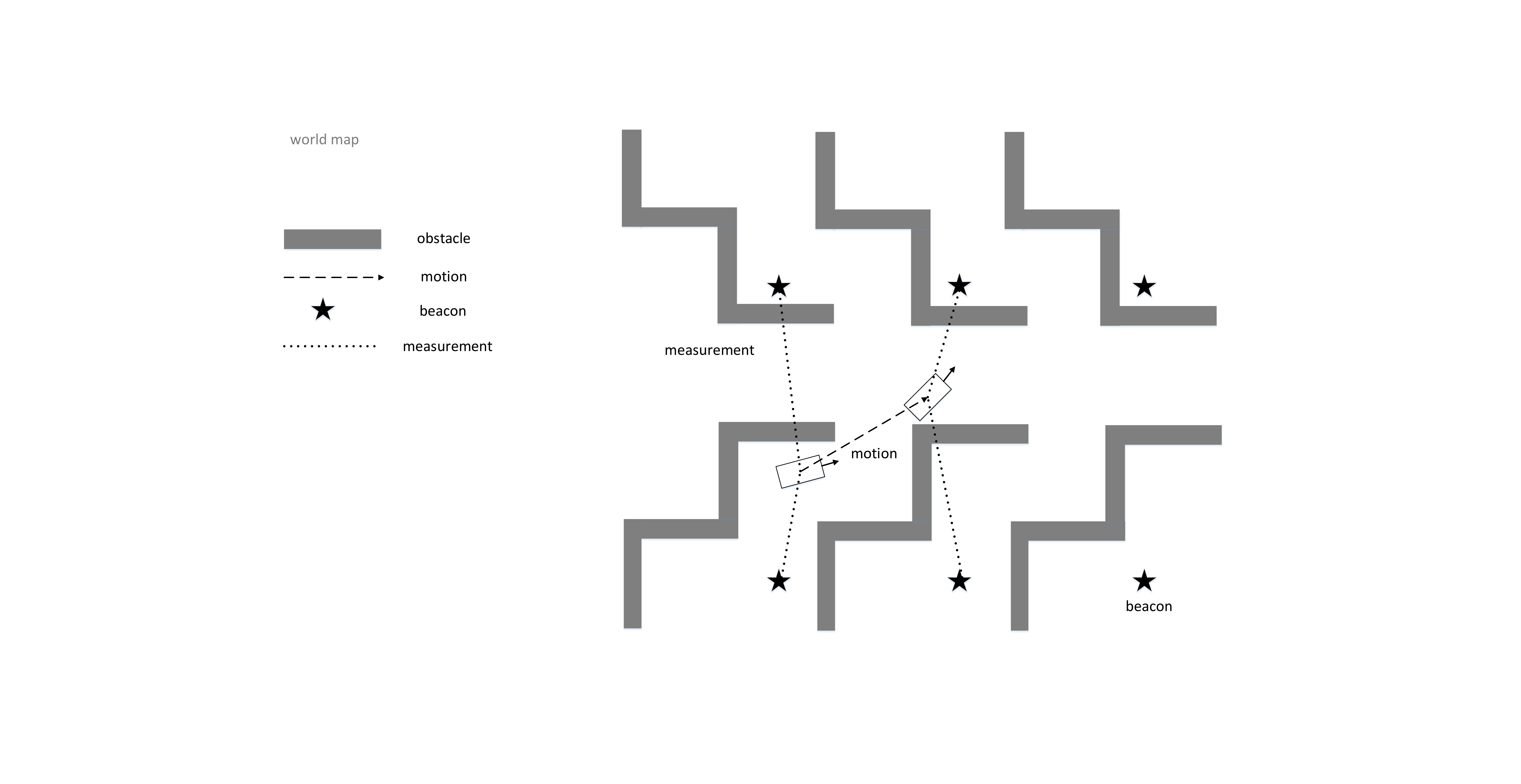}
    \vspace{-1cm}
    \caption{Localization problem visualization. The object moves along the dashed line. The distances from the object to the beacons are measured along the dotted lines. The beacons are marked as stars. The obstacles are shown as  blocks}
    \label{fig::localization}
\end{figure}

Visualization of the object localization problem is shown in Figure~\ref{fig::localization}. 
Here the state $\rvx_t$ consists of coordinates in the plane and heading, $\rvz_t$ are distances between the object and beacons. 
The dashed arrows denote object motion between the states at two consecutive moments of time.
Note that, both state vectors and measurement vectors are noisy.

Assume that we know a motion equation, which relates states at two consecutive moments of time $\rvx_t$ and state $\rvx_{t-1}$:
\begin{equation}
\label{eq::motion}
\rvx_t = f(\rvx_{t - 1}, \rvu_{t}, \boldsymbol{\eta}),
\end{equation}
where $\rvu_{t}$ is an external control and $\boldsymbol{\eta}$ is a motion noise.
In the example from Figure~\ref{fig::localization}, the state is updated as:
 \begin{equation}
 \begin{split}
 &x_{t}=x_{t-1}+(u_t+\eta_r) \cos(\phi_{t-1}+ \Delta \phi_{t} + \eta_{\phi})\\ 
 &y_{t}=y_{t-1}+(u_t+\eta_r) \sin(\phi_{t-1}+\Delta \phi_{t} + \eta_{\phi})\\
 &\phi_{t}=\phi_{t-1} + \Delta \phi_{t} + \eta_{\phi},
 \end{split}
 \end{equation} 
where $\boldsymbol{\eta}=(\eta_r, \eta_{\phi})$ are radial and tangential components of the noise $\boldsymbol{\eta}$, which model uncertainty in object motion.
Also, external control vector $\rvu_t$ has the following form $\rvu_t = [u_t, \Delta \phi_t]$. 

Since there is a noise in the motion equation, we aim to correct the next state with additional measurements that fine-tunes the location of the object in the environment. 
Assume that the measurement equation is known:
\begin{equation}
\label{eq::measurement}
\rvz_t = g(\rvx_t, \boldsymbol{\zeta}),
\end{equation}
where $g: \mathbb{R}^d\rightarrow \mathbb{R}^k$ is a measurement function, which maps $d$-dimensional state $\rvx_t$ to the $k$-dimensional measurement~$\rvz_t$.
Also, denote by $\boldsymbol{\zeta}$ the measurement noise. 
In the aforementioned example (see~Figure~\ref{fig::localization}), the measurement function~$g$ computes the distances between the current object position and the beacons' positions, i.e. $z_k=\|\rvx^{(p)}_t - \rvs_k \| + \zeta_k$, where $\rvs_k$ is the position of the $k$-th beacon and $\rvx^{(p)}_t$ is the position of the object, which is a subvector of state vector $\rvx_t$.

\begin{figure}[!h]
    \centering
    \begin{subfigure}{0.45\textwidth}
    \includegraphics[width=\textwidth]{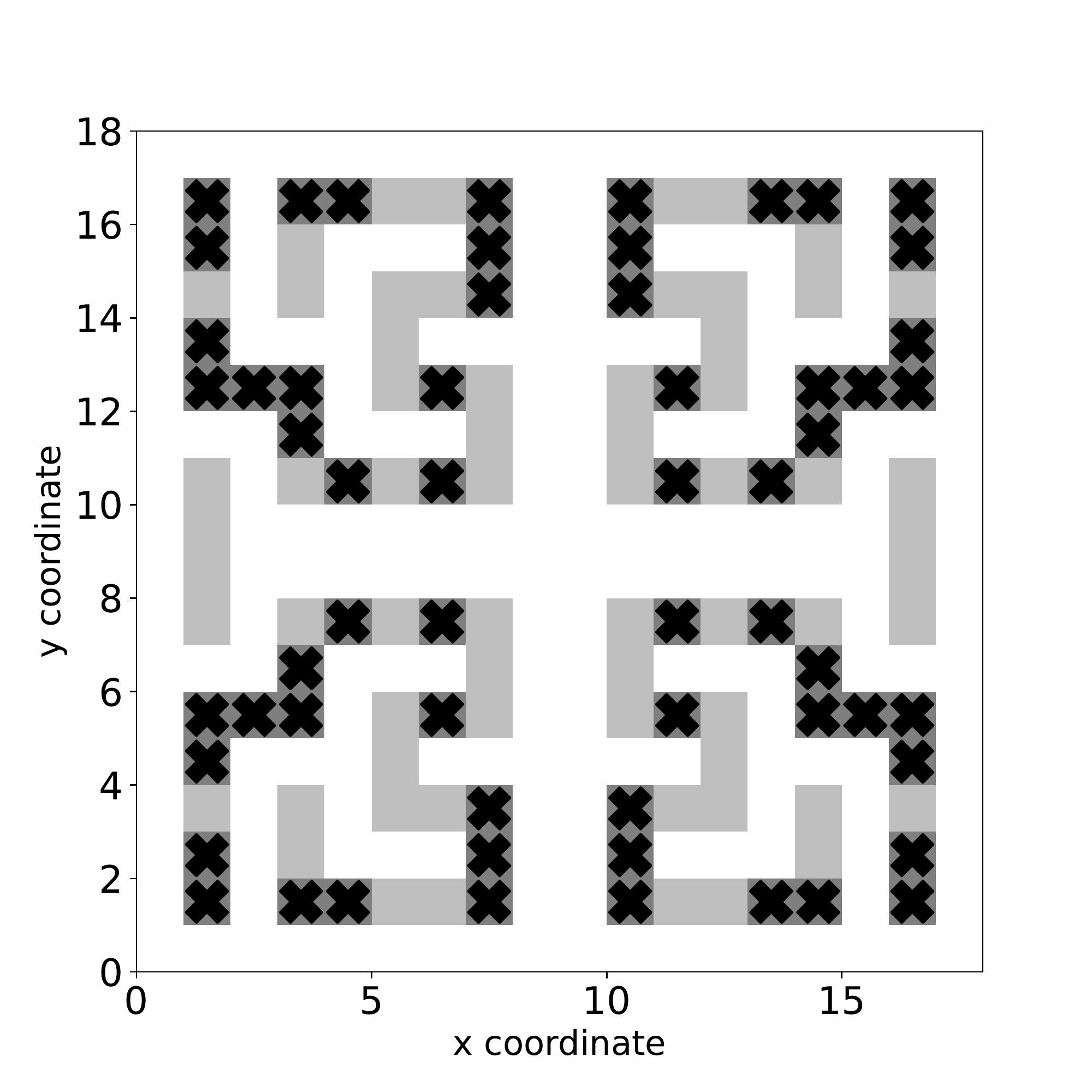}
    \end{subfigure}
    ~
    \begin{subfigure}{0.45\textwidth}
    \includegraphics[width=\textwidth]{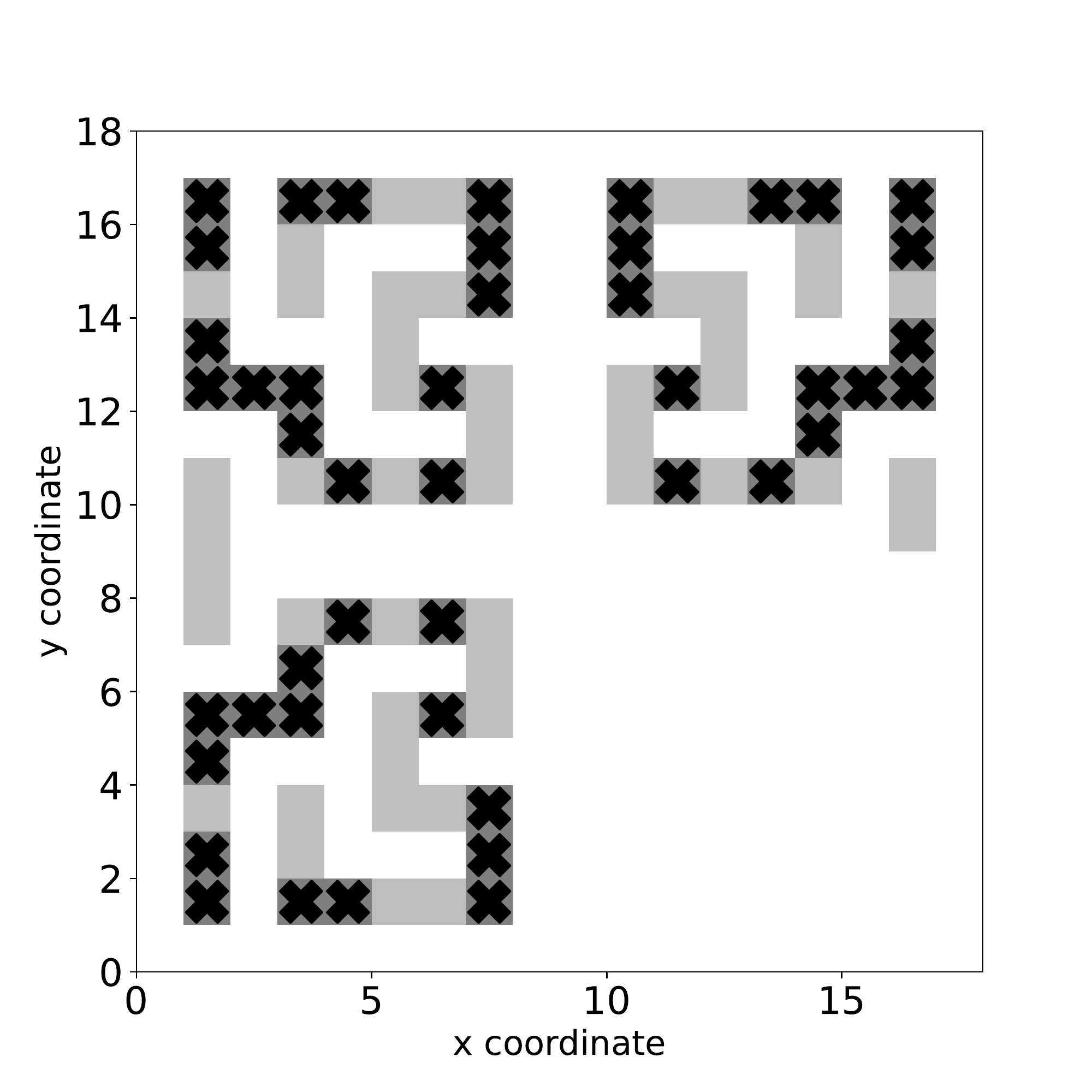}
    \end{subfigure}
    \caption{Example of symmetric (left) and non-symmetric (right) environments. Black crosses indicate beacons. Grey blocks indicate obstacles.}
    \label{fig::sym_env_example}
\end{figure}

Despite the measurements aimed to correct object state, a filtering algorithm may show poor performance if the environment is highly symmetric. 
The environment symmetry may cause that different states $\rvx_t^{(i)}$ are mapped to similar measurements $\rvz_t^{(i)}$. 
If the difference between $\rvz_t^{(i)}$ is within measurement noise, the filter algorithm provides an incorrect state vector. 
Figure~\ref{fig::sym_env_example} shows examples of symmetric and non-symmetric environments.
A more detailed discussion of symmetric environments is presented in Section~\ref{sec::experiments}, where the evaluation of filtering methods in such environments is presented.

The localization problem can be formally stated as the minimization of the mean squared error between the predicted states and the ground-truth ones in the time moments $t=1,\ldots,T$:
\begin{equation}
\begin{split}
&\min_h \frac{1}{T}\sum_{t=1}^{T}{\|h(\rvx_t, \rvz_t)-\rvx^{*}_t\|_2^2},\\
\text{s.t. } & \rvx_t = f(\rvx_{t - 1}, \rvu_{t}, \boldsymbol{\eta}_t)\\
& \rvz_t = g(\rvx_t, \boldsymbol{\zeta}),
\end{split}
\label{eq:problem_statement}
\end{equation}
where $\rvx^{*}_t$ denotes the ground-truth state at time $t$, and function $h$ depends on both state and measurement vectors and provides the estimate of the ground-truth state. 
Further, we use the mean squared error (MSE) loss function: 
\begin{equation}
MSE = \frac{1}{T}\sum_{t=1}^{T}{\|h(\rvx_t, \rvz_t)-\rvx^{*}_t\|_2^2}
\label{eq::mse_def}
\end{equation}
and the final state error (FSE) loss function:
\begin{equation}
FSE = \|h(\rvx_T, \rvz_T)-\rvx^{*}_T\|_2^2
\label{eq::fse_def}
\end{equation}
to evaluate the performance of the considered methods.
In problem~(\ref{eq:problem_statement}) target function $h$ encodes a particular filtering method that eliminates the noise from the measured state, e.g. Kalman filter or particle filter.
A brief description of these filters is given below for the readers' convenience.




\section{Filtering algorithms based on the Kalman filter}

The Kalman filter is based on the assumption of Gaussian distribution of a state, control, and measurements vectors, i.e. $\vx_t\sim\mathcal{N}(\rvx_t, \mP_t)$, $\vu_t\sim \mathcal{N}(\rvu_t, \mQ)$, and $\vz_t\sim\mathcal{N}(\rvz_t, \mR)$, where~$\mP_t$, $\mQ$, and $\mR$ are the state, process, and measurement covariance matrices.
At each step, the state $\rvx_t$ is updated with a linear transformation $\mF$ and the known control dynamics~$\rvu_t$: 
\begin{equation}
    \rvx^-_{t+1} = \mF \rvx_t + \rvu_t.
    \label{eq::Kalman_motion}
\end{equation} 
Besides the measurement vector $\rvz^*_t$, we compute the predicted measurement vector 
\begin{equation}
    \rvz_t=\mH\rvx_t,
    \label{eq::Kalman_pred_measurement}
\end{equation}
which plays the crucial role in the correction of $\rvx^-_{t+1}$.
To correct updated state $\rvx^-_{t+1}$, Kalman filter uses the following equation 
\begin{equation}
    \rvx_{t+1} = \rvx_{t+1}^- + \mK_{t+1}(\rvz^*_{t+1} - \rvz_{t+1}),
    \label{eq::Kalman_x_update}
\end{equation}
where the Kalman gain $\mK_{t+1} = \mP_{t+1}^- \mH^\top (\mH \mP_{t+1}^- \mH^\top + \mR)^{-1}$.
The state covariance matrix~$\mP_{t+1}$ firstly predicted from motion equation as 
\begin{equation}
    \mP_{t+1}^- = \mF \mP_{t} \mF^\top + \mQ,
    \label{eq::Kalman_P_update}
\end{equation}
where $\mQ$ is a pre-defined constant process covariance matrix related to the motion noise $\boldsymbol{\eta}$ from~\eqref{eq::motion}.
And then it is updated with the Kalman gain $\mK_{t+1}$ according to the following formula: $\mP_{t+1} = (\mI - \mK_{t+1} \mH)\mP_{t+1}^-$.
The equations of the state~\eqref{eq::Kalman_motion} and measurement~\eqref{eq::Kalman_pred_measurement} updates are the particular cases of~\eqref{eq::motion} and~\eqref{eq::measurement}, respectively.

This method iteratively gives estimate of state $\rvx_{t+1}$ and covariance matrix $\mP_{t+1}$ from the previous state $\rvx_{t}$ and measurement results $\rvz_{t+1}$. 
The main advantages of Kalman filter are low computational costs and low memory consumption. 
At the same time, it has significant drawbacks, like the linearity of motion and measurement equations, and the assumptions on the normal noise.
 
To generalize the Kalman filter to non-linear motion and measurement equations, the Extended Kalman Filter was proposed~\cite{EKF}. 
It operates with non-linear equations on coordinates $\rvx$ and measurements $\rvz$. 
These non-linear equations are incorporated in the standard Kalman filter as: 
\begin{equation}
\begin{split}
&\rvx_{t+1}^- = f(\rvx_t, \rvu_{t+1}, 0)\\
&\mF_{t+1} = \left.\frac{\partial f(\rvx, \rvu_{t+1})}{\partial \rvx}\right|_{\rvx = \rvx_t}
\end{split}
\qquad \begin{split}
 &\rvz_{t+1} = g(\rvx_{t+1}^-, 0)\\ 
 &\mH_{t+1} = \left.\frac{\partial g(\rvx)}{\partial \rvx}\right|_{\rvx = \rvx^-_{t+1}}\\
 \end{split}
 \label{eq::EKF}
 \end{equation}
In addition, since motion noise $\boldsymbol{\eta}$ is not additive, the covariance process matrix $\mQ \equiv \mQ_t$ depends on time moment  and is recomputed in every time step as follows
\begin{equation}
    \mQ_t = \frac1s \sum_{i=1}^s (\rvx_{t+1}^- - f(\rvx_t, \rvu_{t+1}, \boldsymbol{\eta}_i)) 
 (\rvx_{t+1}^- - f(\rvx_t, \rvu_{t+1}, \boldsymbol{\eta}_i)^\top,
 \label{eq::Qt_EKF}
\end{equation}
where $\boldsymbol{\eta}_i, \; i=1,\ldots,s$ are sampled from some pre-defined distribution $\mathcal{N}(0, \mM)$ corresponding to our assumption on the motion noise.
Other equations in the Extended Kalman filter coincide with equations in the Kalman filter. 


Since the exact computation of Jacobians in~\eqref{eq::EKF} can be computationally intensive,
the unscented Kalman filter~\cite{Julier} computes $f$ and $g$ at the specific sigma points and uses the computed values to approximate the corresponding Jacobians.
This approach reduces the runtime of every iteration but can lead to slow convergence if the approximations of Jacobians are not sufficiently accurate.

In this section, we have briefly described some filtering algorithms inspired by the classical Kalman filter.
They extend the classical Kalman filter to non-linear motion and measurement equations.
However, they still assume that the noise distribution is normal and require the initial object state and its covariance.
These assumptions and requirements limit the practical usage of the aforementioned filtering algorithms in a real-world scenario.

\section{Particle filter algorithm}


As it was previously mentioned, the Kalman filter requires Gaussian distribution of motion and measurement noise and known initial state. 
If these requirements do not hold, the particle filter method~\cite{PF1996} can help.
This method successfully operates with arbitrary  distributions of state and measurement noise and even with the unknown initial state.

The idea of the particle filter method is to generate a set of trial state vectors $\rvx_{t=0}^{i} \in \mathbb{R}^d, \; i=1,\ldots,N$, which are called particles and the  corresponding weights~$w^i$ which are initialized as $1 / N$.
These vectors are used to approximate unknown distribution $p(\vx_t)$ and weights
\begin{equation}
    w^i_t = \mathbb{P}(\vx_{t} = \rvx_t^{i}).
\label{eq::weights}
\end{equation}
Then, the particle states are updated according to the motion equation: $\rvx_{t+1}^i= f(\rvx_t^i, \rvu_t, \boldsymbol{\eta}^i)$, where external control $\rvu_t$ is the same for all particles. 
At this stage, the information is partially lost due to the uncertainty $\boldsymbol{\eta}^i$ in the external control.

After that, we perform the measurement and obtain  $\rvz_{t+1}$.
Then, to estimate the uncertainty in the measured $\rvz_{t+1}$, we compute its conditional likelihood given state vector $\rvx_{t}$: 
\begin{equation}
    \mathcal{L}(\rvz_t|\rvx_t^i) = \prod_{j=1}^K p(\rvz^j_t|\rvx^i_t),
    \label{eq::likelihood}
\end{equation} 
where $K$ is a number of beacons, see Section~\ref{sec::problem_statement}. 
Here $p(\rvz^j_t|\rvx_t^i)$ is the probability to get measurement value $\rvz^j_t$ for beacon with index $j$ at time $t$ given the state $\rvx^i_t$.
In the general case, $p(\rvz|\rvx)$ is calculated from the distribution of the  measurement noise $\boldsymbol{\zeta}$.
In this study, the distribution of measurement noise is Gaussian, i.e. $\boldsymbol{\zeta}\sim\mathcal{N}(0, \mR)$, where $\mR = \sigma^2 \mI$. 
The predicted values of distances to $K$ beacons from the $i$-th particle are $\rvz_t^i = g(\rvx_t^{i}, \boldsymbol{\zeta})$.
At the same time, we perform measurement from the ground-truth object position to the $K$ nearest beacons and get values $z_1^*,\ldots,z_K^*$ stacked in the vector $\rvz_t^* \in \mathbb{R}^K$.
Therefore, we can estimate the likelihood~\eqref{eq::likelihood} with the following formula: 
$\mathcal{L}(\rvz_{t+1}|\rvx_{t+1}^i) = \frac{1}{(2\pi)^{K/2} \sqrt{\det{\mR}}} \exp \left(
-\frac12 (\rvz^*_t - \rvz_t^{i})^\top \mR^{-1} (\rvz^*_t - \rvz_t^{i})
\right)$.

Then, to compute the updated particles' weights $w^i_{t+1}$ we use likelihood~\eqref{eq::likelihood} and current weights: 
\begin{equation}
\label{eq::particle_weights}
w_{t+1}^i \sim \mathcal{L}(\rvz_{t+1}|\rvx_{t+1}^i) w_t^i,
\end{equation}
where $\sim$ indicates equality up to the normalization factor, i.e. $\sum_{i=1}^N w_{t+1}^i = 1$.
From the updated weights $w^i_{t+1}$ the object state $\rvx_{t+1}$ can be estimated as expectation over the generated particles:
\begin{equation}
\begin{split}
& \rvx_{t+1} = \E [\vx_{t+1}] = \sum_{i=1}^N w^i_{t+1} \rvx_{t+1}^i.\\
\end{split}
\end{equation}
According to~\cite{proof} at a large enough number of particles and sufficiently large time steps, values $\rvx_t$ converge to known values $\rvx_t^*$. 

As it was mentioned above, despite the simplicity of implementation and universality of this method, it has drawbacks such as degeneracy and impoverishment~\cite{fight}.
The formal measure of degeneracy is the number of effective particles:
\begin{equation}
    N_{eff} = \left\lfloor\frac{1}{\sum_{i=1}^{N} (w^i)^2}\right\rfloor,
\end{equation}
which varies from $1$ to $N$. 
The worst case is $N_{eff} = 1$, i.e. the single particle has non-zero weight. 
The best case is $N_{eff} = N$, i.e. all particles have the same values of weights $w^i = 1/N$. 
If $N_{eff}$ drops below a pre-defined threshold, for example, $N_{eff} \le N / 4$ or $N_{eff} \le  N / 2$, it indicates a degeneracy problem.
To address this problem, a resampling procedure is used.

The widespread resampling procedure is known as stochastic resampling~\cite{resamp1, Tutorial}. 
Formally it samples $N$ indices of particles from the multinomial distribution with repetitions.
The parameters of multinomial distributions are weights $w_t^i$.
After that, the $j$-th particle state is updated with the $i_j$ particle state, where $i_j$ is sampled index.
The described stochastic resampling procedure is summarized as follows:
\begin{equation}
\begin{split}
    &i_1,\ldots,i_N \sim Multinomial(w^1_{t+1},\ldots,w^N_{t+1})\\
    &\rvx^1_{t+1},\ldots,\rvx^N_{t+1} \leftarrow  \rvx^{i_1}_{t+1},\ldots,\rvx^{i_N}_{t+1}\\
    &w^i_{t+1}=\frac{1}{N}.
\end{split}
\label{eq::stoch_resampling}
\end{equation}

After the resampling of particles is done, few particles have the same states. 
Therefore, they represent the target probability density function poorly and particle filter may converge to the wrong state.
This problem is known as impoverishment~\cite{fight}. 
To improve the diversity of particles, random noise with sufficiently large variance is added to particle states~\cite{towards}.
The described particle filter method is summarized in Algorithm~\ref{alg::PF}.


\begin{algorithm}[!h]
\caption{Particle filter method with stochastic resampling.}
\label{alg::PF}
\begin{algorithmic}[1]
\REQUIRE Number of particles $N \geq 1$, number of beacons $K \geq 1$, motion and measurement equations $f$, $g$, covariance matrices $\mM$ and $\mR$ of motion and measurement noise.
\ENSURE predicted object states $\rvx_{t}$ for $t=1,\ldots,T$
\FOR{$i = 1$ to $N$}
    \STATE Initialize weights $w^i \gets 1/N$
    \STATE Initialize particle state $\rvx_1^i \gets \mathcal{U}(\rvx_{\min},\rvx_{\max})$
 \ENDFOR
\FOR{$t=1$ to $T$}
    \FOR{$i=1$ to $N$}
        \STATE Generate $\boldsymbol{\eta}^i \sim \mathcal{N}(0, \mM)$, $\boldsymbol{\zeta}^i \sim \mathcal{N}(0, \mR)$
        \STATE Update state $\rvx_{t+1}^i = f(\rvx_t^i,\rvu_t, \boldsymbol{\eta}^i)$ and perform measurements $\rvz^i_{t+1} = g(\rvx_{t+1}^i, \boldsymbol{\zeta}^i)$
        \STATE Compute likelihood $\mathcal{L}(\rvz_{t+1}|\rvx_{t+1}^i) = \frac{1}{(2\pi)^{K/2} \sqrt{\det{\mR}}}\exp \left( -\frac12 (\rvz^*_t - \rvz_t^{i})^\top \mR^{-1} (\rvz^*_t - \rvz_t^{i}) \right)$
        \STATE Update particle's weight $w_{t+1}^i \sim \mathcal{L}(\rvz_{t+1}|\rvx_{t+1}^i) w_t^i$
    \ENDFOR
    \STATE Perform resampling of the updated states according to~\eqref{eq::stoch_resampling}
    \STATE $\rvx_{t+1} \gets \sum_{i=1}^N w_{t+1}^i \rvx_{t+1}^i$
\ENDFOR
\end{algorithmic}
\end{algorithm}

\begin{figure}[!h]
    \centering
\includegraphics[width=0.5\textwidth]
    {{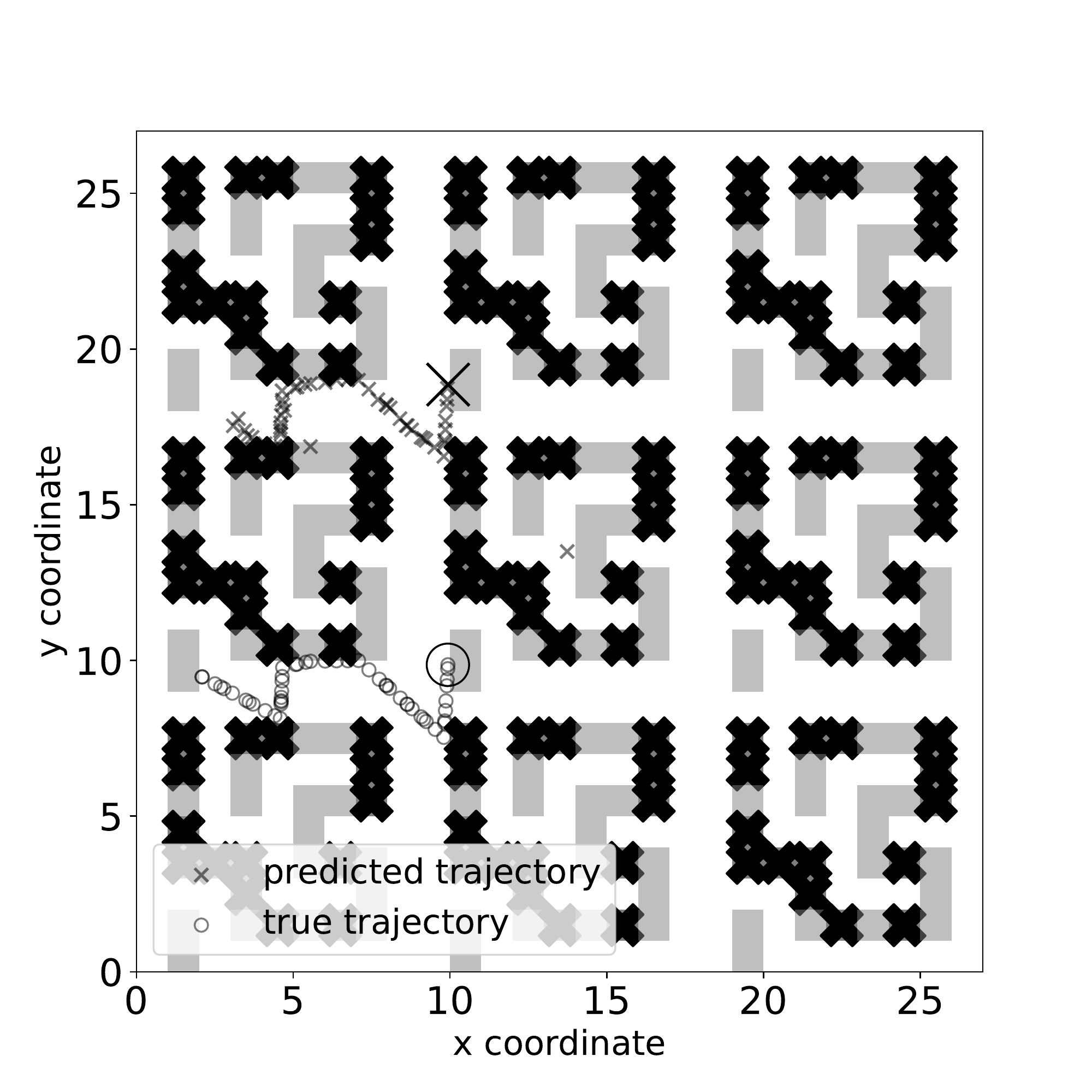}}
    \caption{Visualization of the symmetric test environment with the ground truth and filtered trajectories marked with circles and crosses, respectively. The measurements are performed from the five nearest beacons. The final positions are shown with a big circle and cross. Note that the final filtered points may differ from the corresponding points in the ground-truth trajectory, though they are identical up to symmetry. Number of particles $N=10000$.}
    \label{fig::symmetric_env_traj}
\end{figure}

The impoverishment problem might be especially severe in highly symmetric environments. 
In this case, consistent state representation in several similar subparts of the environment requires the number of particles proportional to the number of subparts, see Figure~\ref{fig::sym_env_example}~(left). 
Moreover, if the environment is highly symmetric, the filtered trajectory may coincide with the ground truth but only up to the symmetric subpart, see Figure~\ref{fig::symmetric_env_traj}, where the predicted trajectory is computed by the particle filter method.
To reduce the number of particles necessary for convergence in a symmetric environment, we suggest equipping every particle with equations~\eqref{eq::EKF} and~\eqref{eq::Qt_EKF} from the Extended Kalman filter. 
A detailed description of the proposed Multiparticle Kalman filter (MKF) is given in the next section.

\section{Multiparticle Kalman filter}

We propose a new natural combination of Kalman and particle filters. 
The goal of such a combination is to develop an algorithm, which deals with the unknown initial states like particle filter and accelerates convergence to an optimal state like the Kalman filter.
Since the symmetric environments are especially challenging for the particle filter, we will use them to illustrate the accelerated convergence of the proposed method.

The idea of the proposed method is to generate a set of trial state vectors (particles) $\rvx_{t=0}^{i} \in \mathbb{R}^d, \; i=1,\ldots,N$ with corresponding weights~$w_{t=0}^i = 1/N$, and also covariance matrices $\mP_{t=0}^i$.
Again, vectors $\rvx_t^{i}$ and weights $w_t^i$ play the same role as in the particle filter.

Then, we use the formulas from the Extended Kalman filter for every generated particle in parallel. 
In particular, equations~\eqref{eq::Kalman_x_update},~\eqref{eq::Kalman_P_update},~\eqref{eq::EKF},~\eqref{eq::Qt_EKF} are used to compute the updated state vector of every particle $\rvx^i_{t+1}$.
After that, to update weights $w_t^{i+1}$ we compute distances to the beacons from the updated states $\rvz_{t+1}^{i} = g(\rvx_{t+1}^{i+})$.
Then the weights are updated in the same way as in the particle filter based on the likelihood 
\[
\mathcal{L}(\rvz_{t+1}|\rvx_{t+1}^{i+}) = \prod_{j=1}^K p(\rvz^j_{t+1}|\rvx^{i+}_{t+1}),
\]
where $K$ is the number of beacons, which are used to measure distances.
Now, the updated weights are computed as follows:
\begin{equation}
w_{t+1}^i \sim \mathcal{L}^i(\rvz_{t+1}|\rvx_{t+1}^{i+}) w_t^i.
\end{equation}

To prevent the degeneracy phenomenon, we perform resampling of the obtained particle states.
We follow the resampling procedure used in the particle filter~\eqref{eq::stoch_resampling} and modify it to resample not only particle states $\rvx^{i}_t$ but also corresponding covariance matrices $\mP^i_{t+1}$. 
The modified resampling procedure used in the proposed method is summarized in~\eqref{eq::kpf_resampling}:
\begin{equation}
\begin{split}
    &i_1,\ldots,i_N \sim Multinomial(w^1_{t+1},\ldots,w^N_{t+1})\\
    &\rvx^{1+}_{t+1},\ldots,\rvx^{N+}_{t+1} \leftarrow  \rvx^{i_1+}_{t+1},\ldots,\rvx^{i_N+}_{t+1}\\
    &\mP^{1}_{t+1},\ldots,\mP^{N}_{t+1} \leftarrow  \mP^{i_1+}_{t+1},\ldots,\mP^{i_N+}_{t+1}\\
    &w^i_{t+1}=\frac{1}{N}.
\end{split}
\label{eq::kpf_resampling}
\end{equation}
Now to address the impoverishment issue, we add noise to the resampled states: 
\begin{equation}
    \rvx^{i}_{t+1} = f(\rvx^{i+}_{t+1}, \boldsymbol{0}, \boldsymbol{\eta}^i),
\end{equation}
where $\boldsymbol{\eta}^i \sim \mathcal{N}(0, \mM)$, where $\mM$ is given covariance matrix.


Note that the per-iteration computational complexity of the presented algorithm is higher than the complexity of the standard particle filter. 
Indeed, in addition to the particle filter steps, the proposed method processes $d \times d$ matrices for every particle.
Therefore, to process every particle memory complexity is increased up to $O(d^2)$ caused by storing matrices, and computational complexity is increased up to $O(d^3)$ operations caused by matrix multiplications\footnote{This complexity can be slightly reduced to $O(d^{2.32})$ according to~\cite{duan2022faster}}. 
Despite this, we observe in the experiments (see Section~\ref{sec::experiments}) that it ensures faster convergence since requires fewer particles.

\section{Computational experiments}
\label{sec::experiments}

In this section, we present the description of the experiments for comparison of the proposed multiparticle Kalman filter (MKF) with a standard particle filter in symmetric and non-symmetric environments.
We exclude the Kalman-based filters from our comparison since they require the object's initial state, which is unknown according to our assumption.
Every experiment is conducted on a single NVIDIA Tesla V100 GPU.

\subsection{Test environments}

\begin{figure}[!ht]
\centering
    \begin{subfigure}{0.3\textwidth}
    \centering
    \includegraphics[width=\textwidth]
    {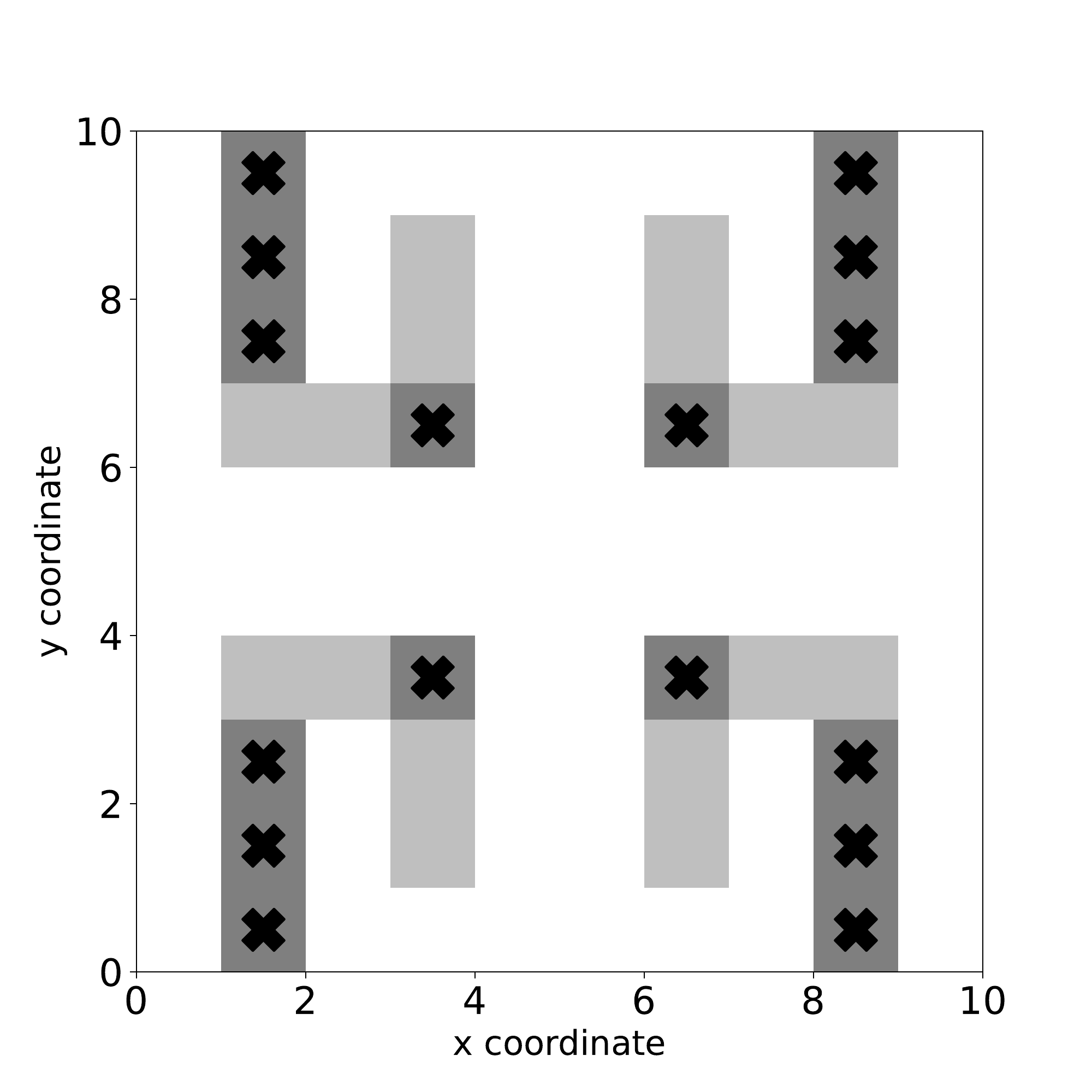}
    \caption{\textit{world $10\times 10$}}
    \end{subfigure}
    ~
    \begin{subfigure}{0.3\textwidth}
    \centering
    \includegraphics[width=\textwidth]
    {pics/_w18.pdf}
    \caption{\textit{World $18\times 18$}}
    \end{subfigure}
    ~
    \begin{subfigure}{0.3\textwidth}
    \centering
    \includegraphics[width=\textwidth]
    {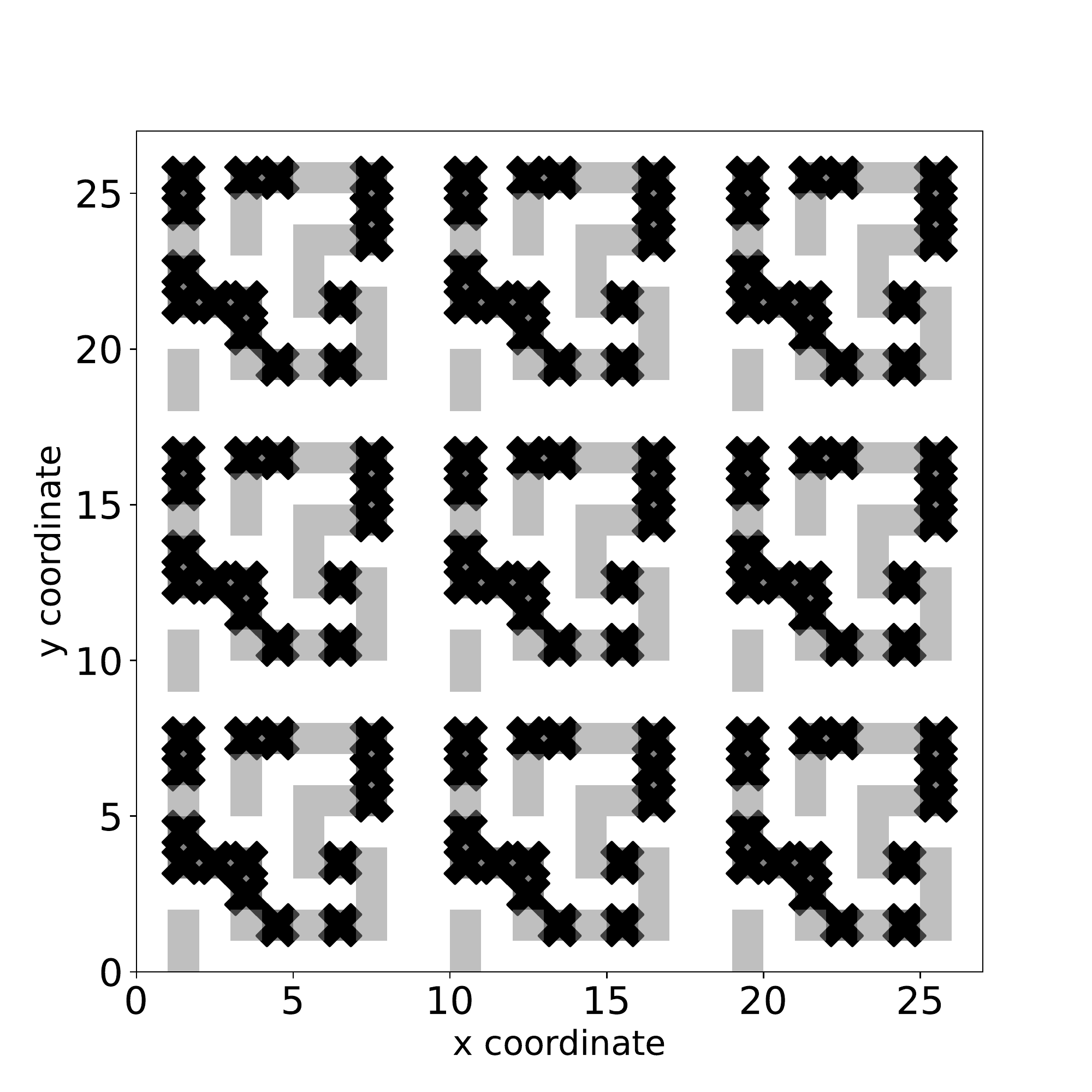}
    \caption{\textit{WORLD $27\times 27$}}
    \end{subfigure}
\caption{Visualization of symmetrical test environments. Beacons and obstacles are shown as black crosses and grey blocks, respectively.}
\label{fig::symmetric_envs}
\end{figure}

To evaluate the performance of the proposed method and compare it with the particle filter we use different types of environments.
In particular, we consider the symmetric environments with an increasing number of symmetrical subparts and call them \textit{world $10\times 10$}, \textit{World $18\times 18$}, and \textit{WORLD $27\times 27$}.
The number of beacons is also increased with the number of symmetrical subparts which makes filtering of object states more challenging.
The considered symmetric environments are shown in Figure~\ref{fig::symmetric_envs}.

\begin{figure}[!ht]
\centering
    \begin{subfigure}{0.3\textwidth}
    \centering
    \includegraphics[width=\textwidth]
    {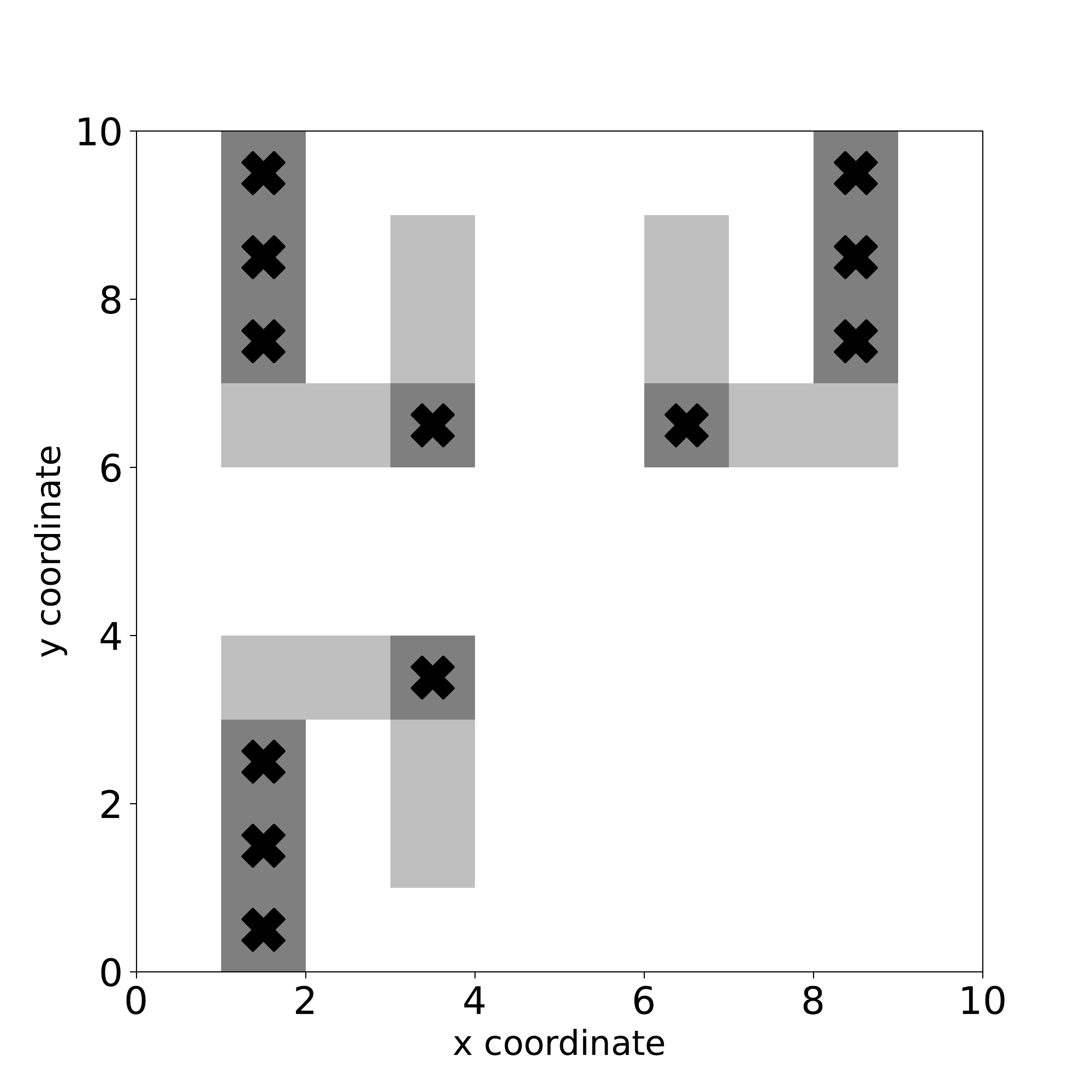}
    \caption{\textit{n-world $10\times 10$}}
    \end{subfigure}
    ~
    \begin{subfigure}{0.3\textwidth}
    \centering
    \includegraphics[width=\textwidth]
    {pics/_w18_1.pdf}
    \caption{\textit{n-World $18\times 18$}}
    \end{subfigure}
    ~
    \begin{subfigure}{0.3\textwidth}
    \centering
    \includegraphics[width=\textwidth]
    {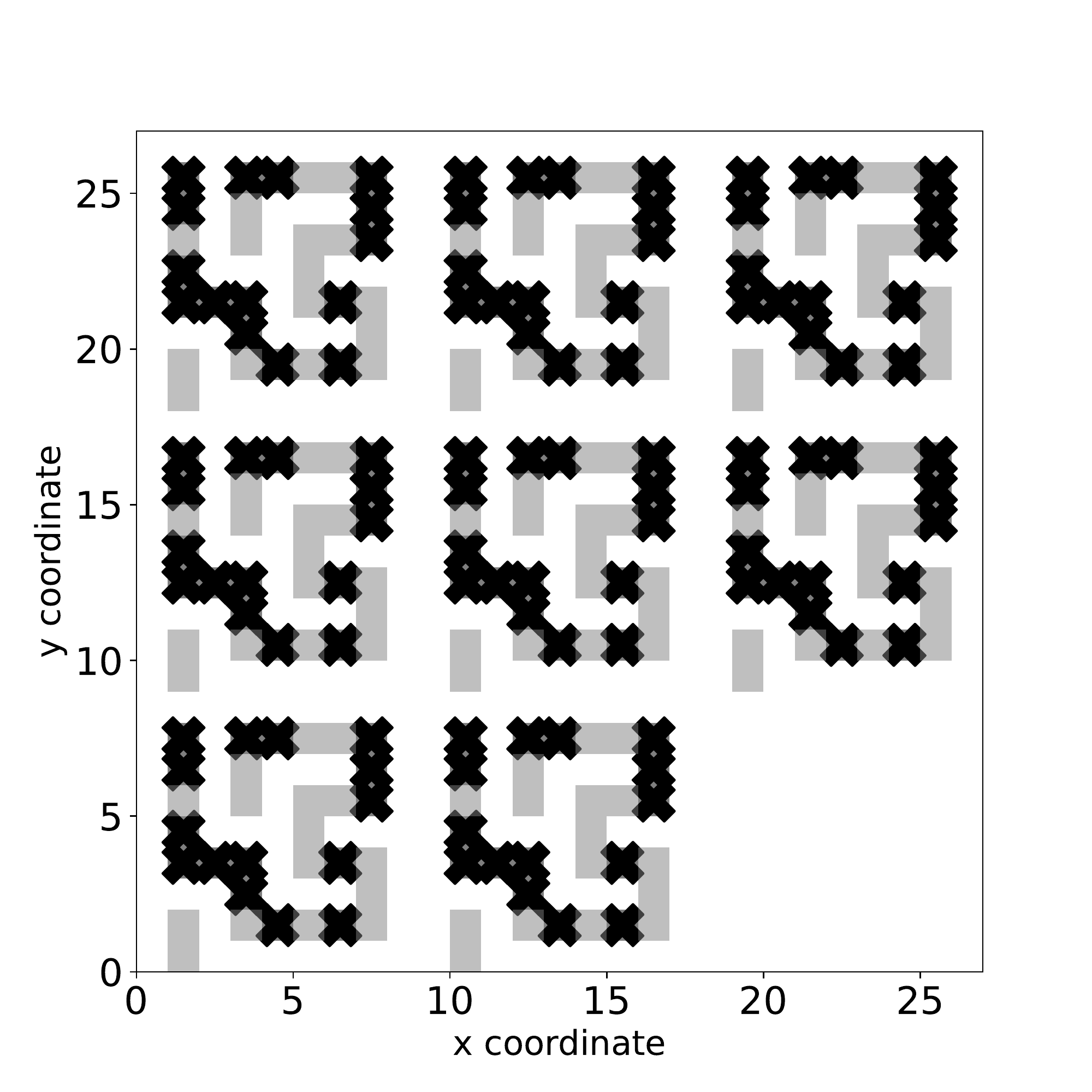}
    \caption{\textit{n-WORLD $27\times 27$}}
    \end{subfigure}
\caption{Visualization of \emph{nonsymmetric} test environments. Beacons and obstacles are shown as black crosses and grey blocks, respectively.}
\label{fig::nonsym_envs}
\end{figure}

To illustrate the effect of symmetry in an environment, we remove a subpart in every environment described above such that they become nonsymmetric.
The nonsymmetric analogs of the aforementioned symmetrical environments are shown in Figure~\ref{fig::nonsym_envs} and we call them \textit{n-world $10\times 10$}, \textit{n-World $18\times 18$}, and \textit{n-WORLD $27\times 27$}, respectively.

\begin{figure}[!ht]
\centering
\includegraphics[width=10cm]{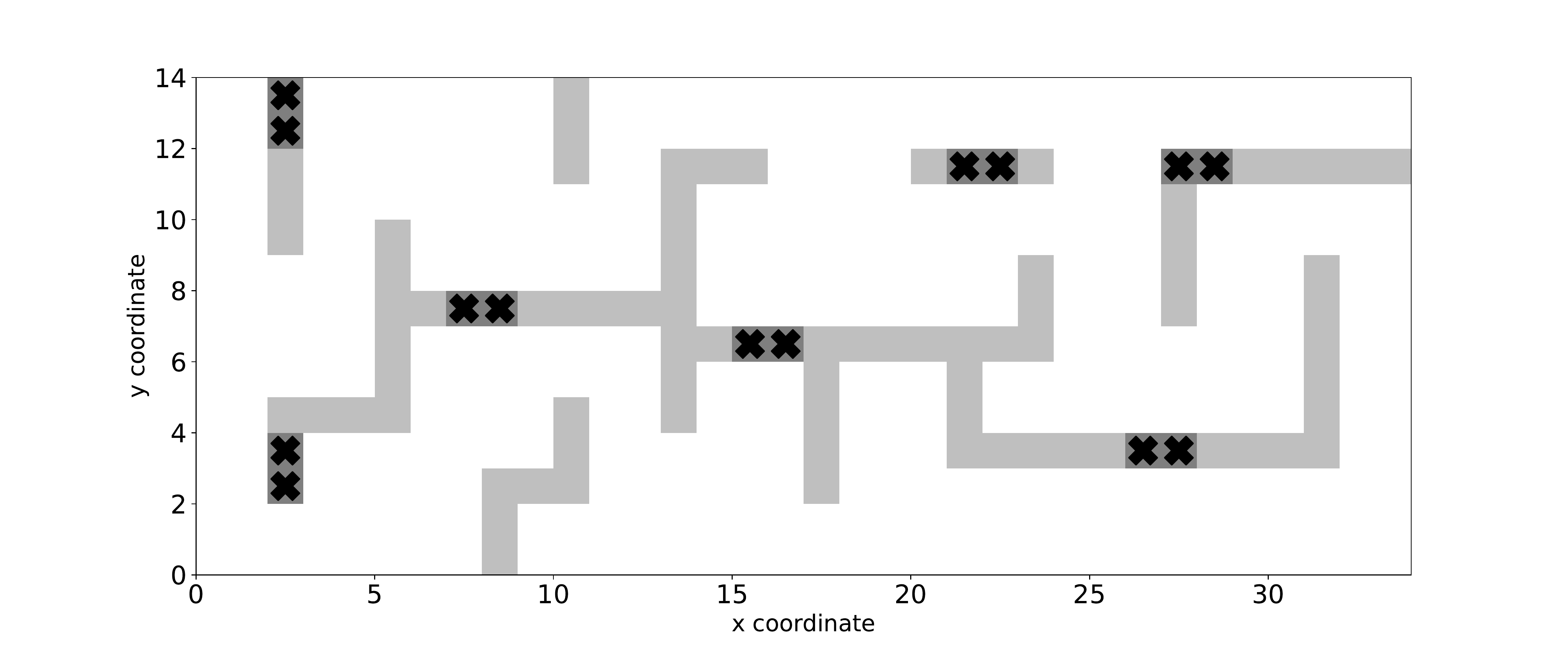}
\caption{Visualization of the \emph{Labyrinth} environment. Beacons and obstacles are shown as black crosses and grey blocks, respectively.
}
\label{fig::labyrinth}
\end{figure}

One more test environment \emph{Labyrinth} is considered in~\cite{towards} and is shown in Figure~\ref{fig::labyrinth}. 
Compared to the previous environments, \emph{Labyrinth} environment has a lower degree of symmetry and allows solving localization problems accurately and comparatively fast in terms of the number of time steps for the convergence.
Therefore, we also compare the considered methods in this relatively friendly environment. 

\subsection{Trajectory generation procedure}

To generate trajectories for an object in the considered environments, the following procedure is used.
The object starts motion from a random location without obstacles and has a randomly chosen direction.
In every step, it moves according to external and known velocity $u \in [0, 0.5]$, and the direction remains the same from the previous step within the noise.
If this movement leads to a collision with an obstacle, the object's direction is changed randomly to avoid collision with another obstacle.
To simulate engine noise, the velocity~$u$ is perturbed by $\eta_r \sim \mathcal{U}[-0.02, 0.02]$ and the direction $\phi$ is also perturbed by $\eta_{\phi} \sim 2\pi \alpha$, where $\alpha \sim \mathcal{U}[-0.01, 0.01]$.
The direction perturbation simulates the uncertainty in the object control system.
In the considered environments, we set the number of time steps in every trajectory $T=100$. 
To make a fair comparison of the considered methods, we generate $10000$ trajectories for testing.

\paragraph{Filter input data.}
The input data for every filter algorithm consists of the following parts: ground-truth measurement vector $\rvz^*_t$, external control vector $\rvu_t = [u_{r}, \Delta \phi_t]$.
An element $\Delta \phi_t \neq 0$ only if the collision with obstacle appears.
Note that, the object movement is affected by the additional noise $\eta_{u}$ and $\eta_{\phi}$.
Therefore, the filtering methods have to identify the ground-truth object state including its position and heading.



\subsection{Hyperparameters}
\label{sec:hyperparams}
Before one runs the proposed filtering method, the following hyperparameters have to be set: covariance matrices of motion and measurement noise $\mM$ and $\mR$, and initial state covariance matrix $\mP_{t=0}$. 
These hyperparameters significantly affect the performance of the method and have to be tuned carefully.
According to~\cite{towards}, a filtering approach based on particles works better if the variances of motion and measurement noise exceed the ground-truth variances in measurement devices and the object control system.
However, the excessively large variance of motion and measurement noise may lead to slow convergence.
In the experiments, we use the following ground-truth variances in measurement devices and the object control system: $\mR_0 = \sigma_{d0}^2 \mI$, where $\sigma_{d0}^2 = 0.01$ and $\mM_0 = \mathrm{diag}(\sigma^2_{r0}, \sigma^2_{\phi 0})$, $\sigma_{r0} = 0.02$, $\sigma_{\phi 0} = 0.01 \cdot 2\pi$.
At the same time, to study the robustness of the proposed approach to different scales of motion and measurement variance, we consider the following settings.
The first setup is $\mM = \mM_0$ and $\mR = 2\mR_0$.
The second setup is $\mM = 4\mM_0$ and $\mR = 2\mR_0$.
The initial state covariance matrix $\mP_{t=0} = \mathrm{diag}(\sigma_{x}^2, \sigma_{y}^2, \sigma_{\phi}^2)$ and $\sigma_{x}^2=\sigma_{y}^2=\frac{w\,h}{12}$ and $\sigma_{\phi}^2=\frac{(2\pi)^2}{12}$, where $w, h$ are width and height of environment and factor $1/12$ is used to model the uniform distribution of particle states in the environment.

\subsection{Comparison of multiparticle Kalman filter with particle filter}

In this section, we provide a comparison of the considered filtering methods in the aforementioned environments.
However, before presenting the comparison results we introduce the upper bound of the MSE error that indicates the poor quality of the state estimate.
The na\"ive filtering method just generates uniformly random states of the object in the given environment.
Therefore, we can estimate MSE between randomly generated states and the ground-truth states for the considered environments as $MSE_{random}=\frac{w^2+h^2}{6}$, where~$w$ and~$h$ are the width and height of the environment, respectively.
If a filtering method generates states such that MSE between them and the ground-truth states is larger than $MSE_{random}$, we consider such filtering completely useless and show this threshold in the plots below. 


In the experiments, we compare the proposed filtering method with the classical particle filter (see Algorithm~\ref{alg::PF}).
Kalman filter and its modifications are excluded from the comparison since they do not perform well without knowledge of the initial state. 
Moreover, the Gaussian distribution of state vectors assumes an elliptical uncertainty region that is irrelevant to the considered environments.
We use both MSE~\eqref{eq::mse_def} and FSE~\eqref{eq::fse_def} loss functions.
Also, we compare the robustness of the considered methods to the scale of motion covariance matrix $\mM$.
In particular, the first setting is $\mM = \mM_0$, which is further referred to as $\Sigma$ in legends.
The second setting is $\mM = 4\mM_0$, which is further referred to as $4\Sigma$ in legends.
Here we denote by $\mM_0$ the ground-truth covariance matrix of the noise from the object control system.
The measurement noise covariance matrix in both settings is $\mR = 2\mR_0$, where $\mR_0$ is the covariance matrix of the noise from a measurement device.
The values for $\mM_0$ and $\mR_0$ used in our simulations are given in Section~\ref{sec:hyperparams}.

\begin{figure}[!ht]
    \begin{subfigure}{0.3\textwidth}
    \includegraphics[width=\textwidth]
    {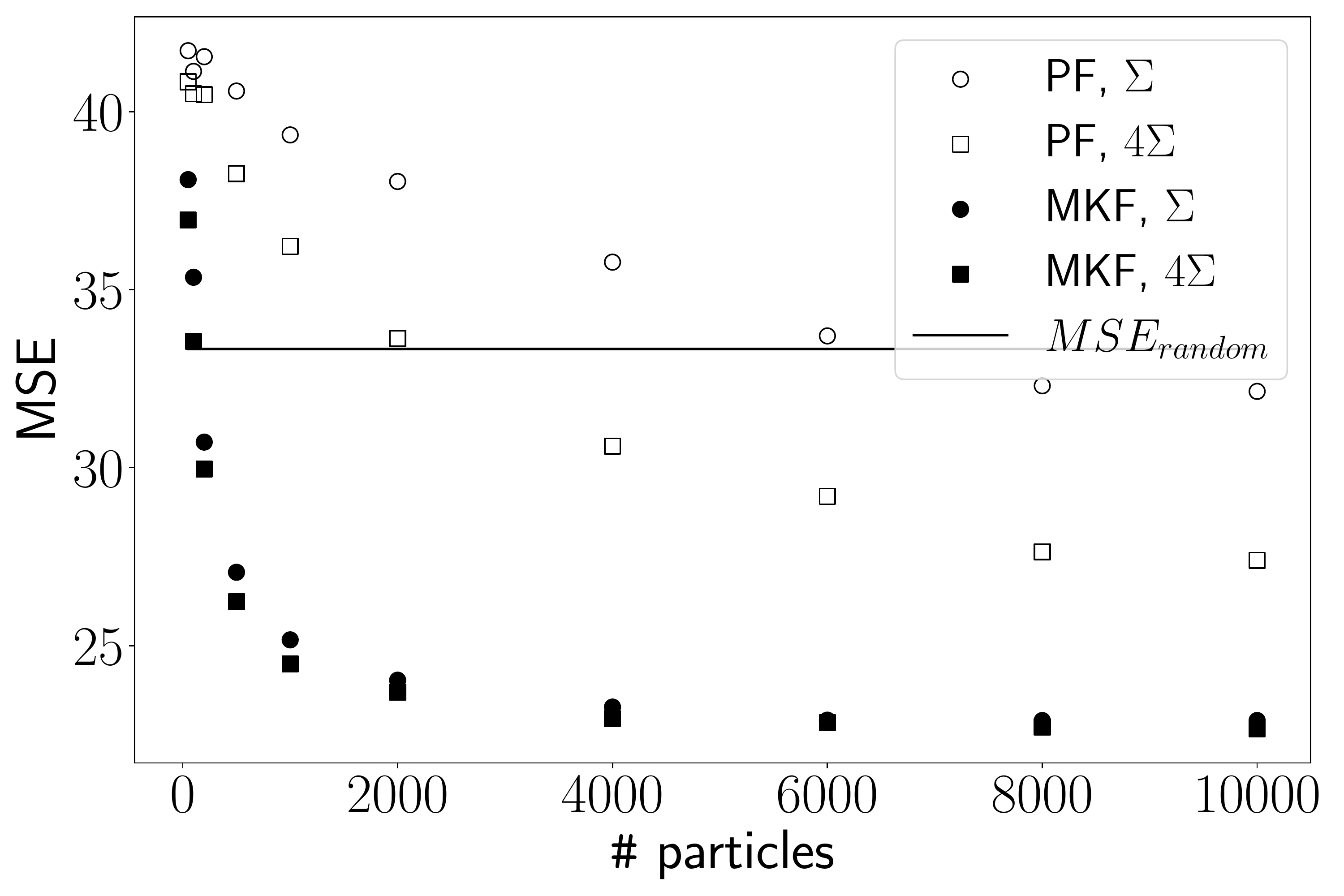}
    \caption{\textit{world $10 \times 10$}}
    \end{subfigure}
    ~
    \begin{subfigure}{0.3\textwidth}
    \includegraphics[width=\textwidth]
    {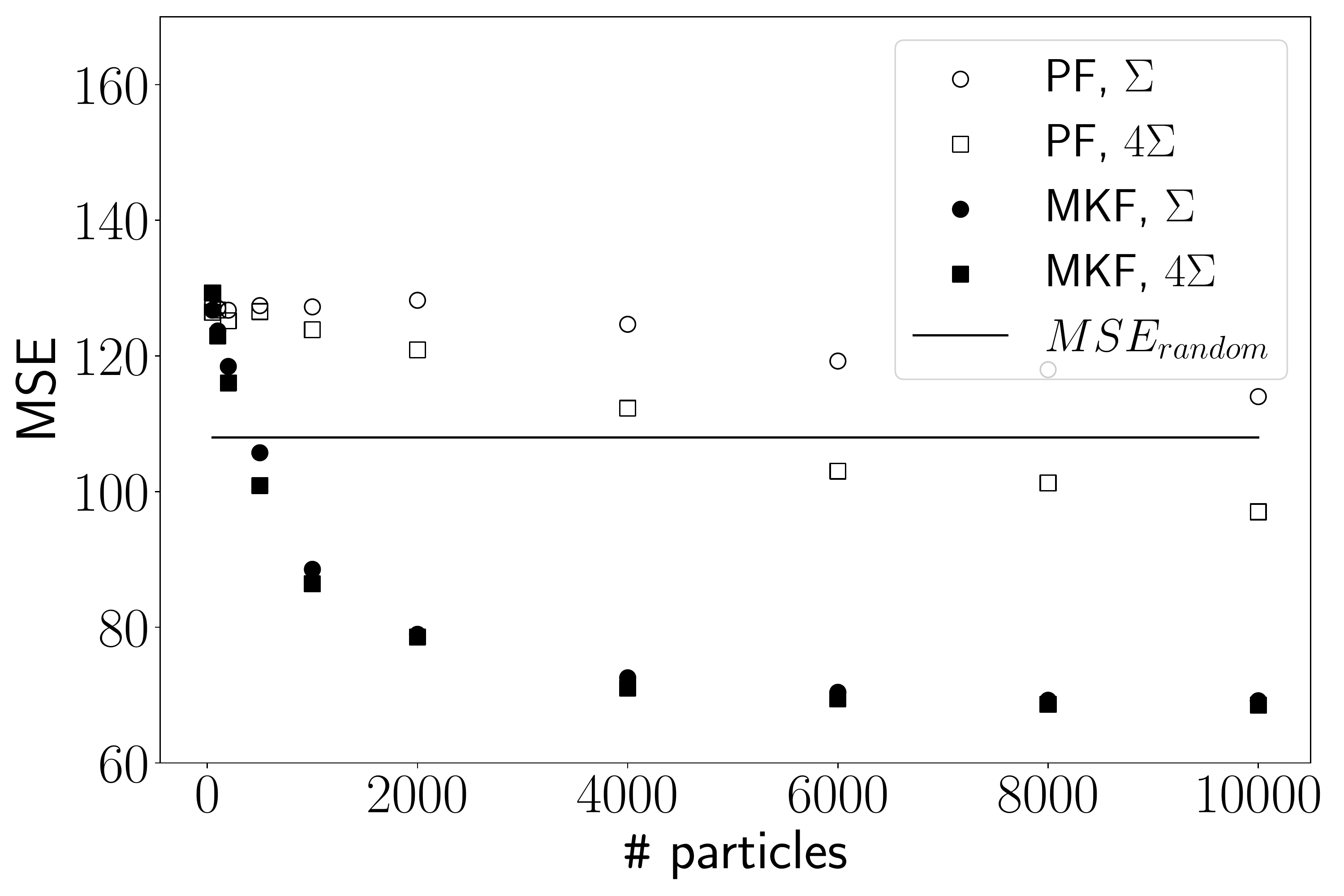}
    \caption{\textit{World $18 \times 18$}}
    \end{subfigure}
    ~
    \begin{subfigure}{0.3\textwidth}
    \centering
    \includegraphics[width=\textwidth]
    {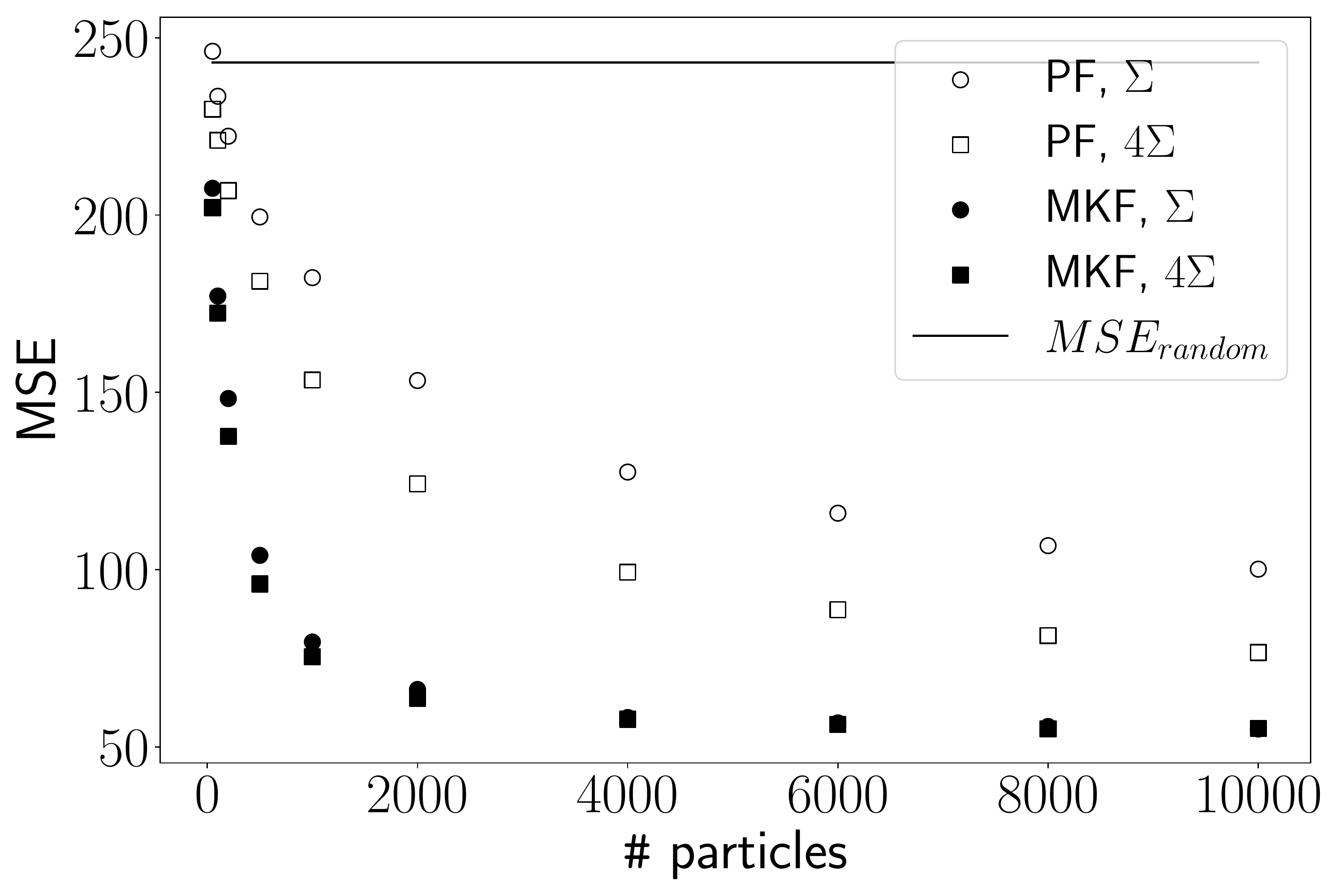}
    \caption{\textit{WORLD $27 \times 27$}}
    \end{subfigure}
\caption{Dependence of MSE on the number of particles in three symmetric environments. Multiparticle Kalman filter (MKF) demonstrates more accurate filtering of states and requires fewer particles for MSE convergence compared to the particle filter (PF). Our method is also less sensitive to the estimate of the motion noise than the particle filter.}
\label{fig::MSE}
\end{figure}

\begin{figure}[!ht]
    \begin{subfigure}{0.3\textwidth}
    \includegraphics[width=\textwidth]
    {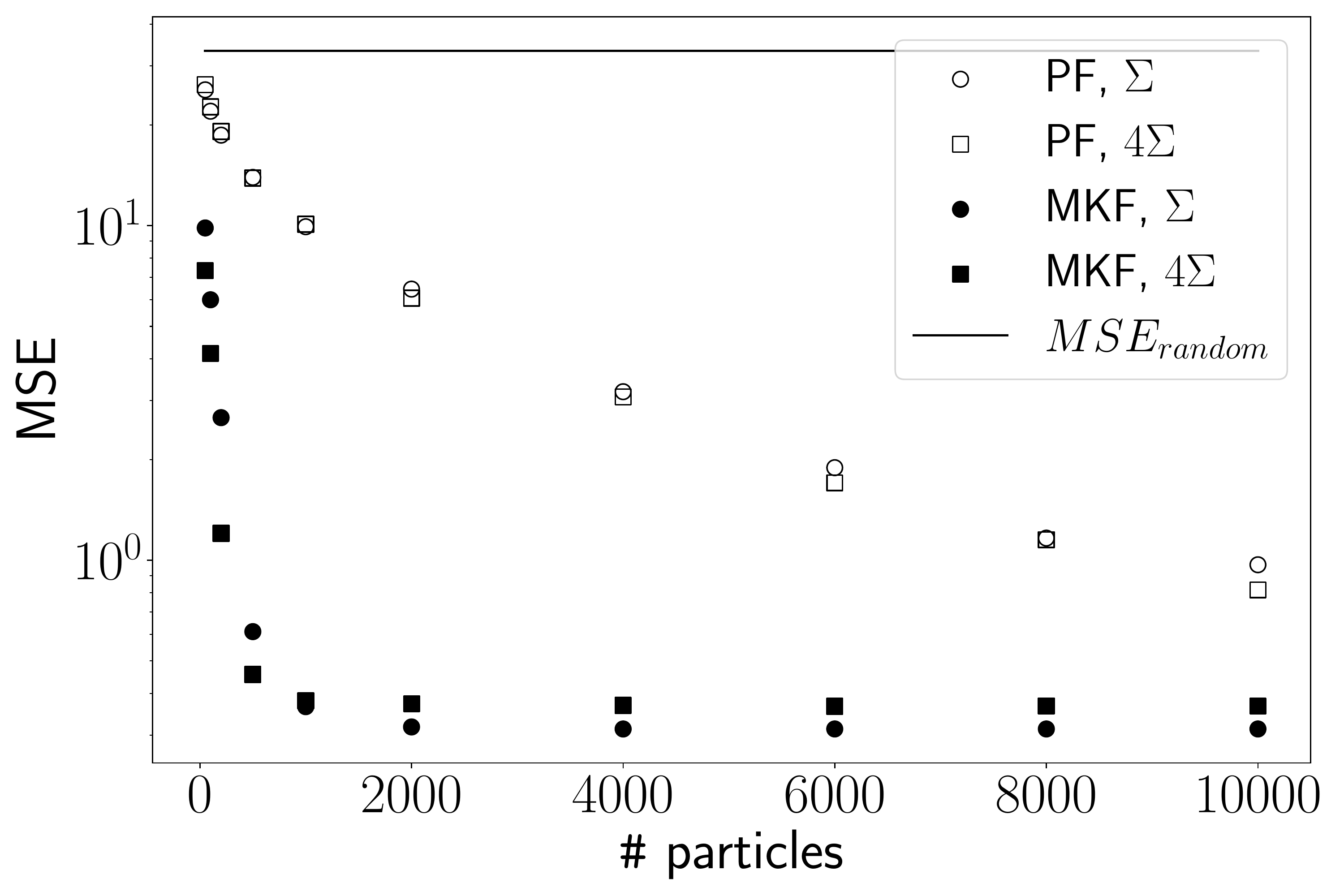}
    \caption{\textit{n-world $10 \times 10$}}
    \end{subfigure}
    ~
    \begin{subfigure}{0.3\textwidth}
    \includegraphics[width=\textwidth]
    {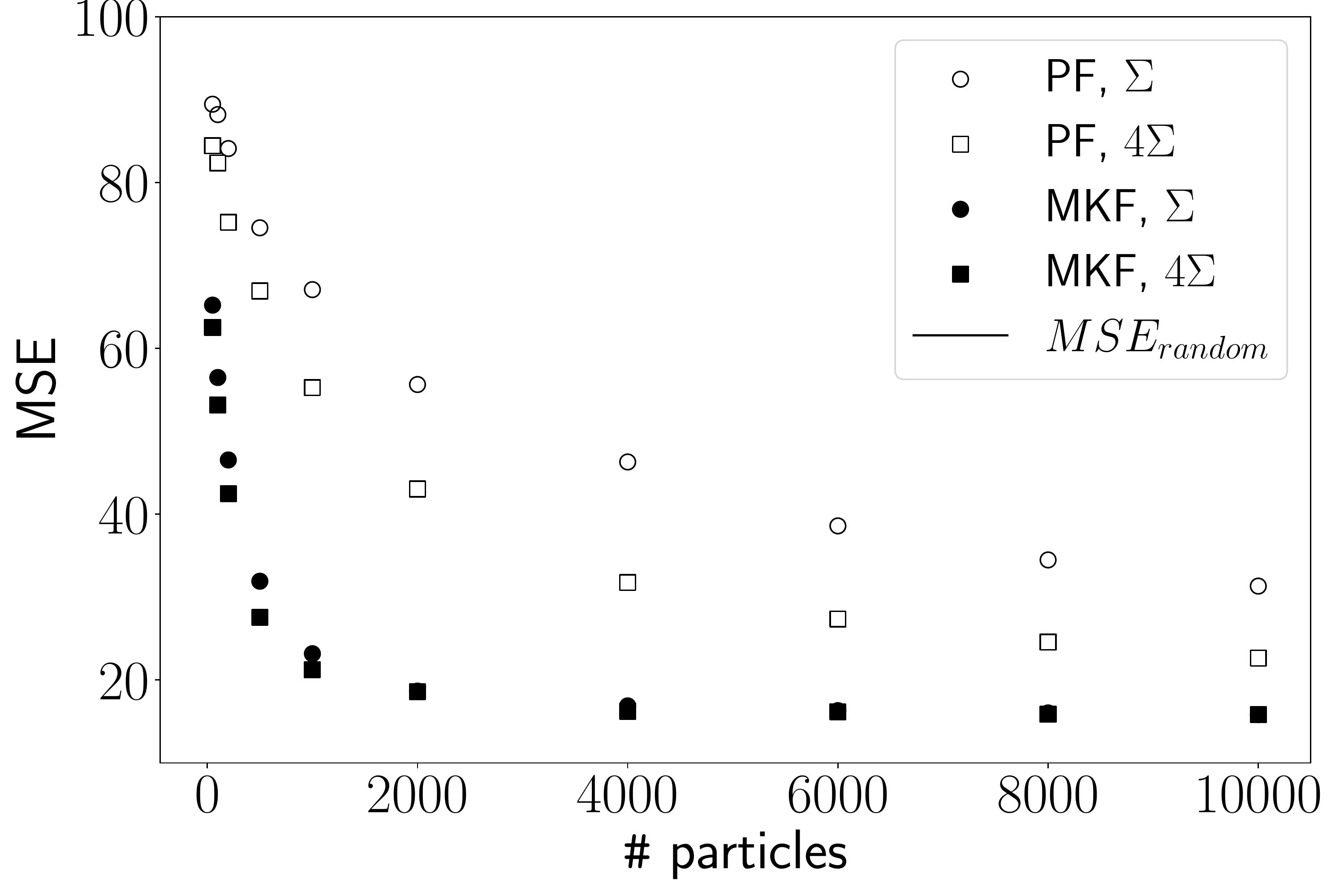}
    \caption{\textit{n-World $18 \times 18$}}
    \end{subfigure}
~
    \begin{subfigure}{0.3\textwidth}
    \centering
    \includegraphics[width=\textwidth]
    {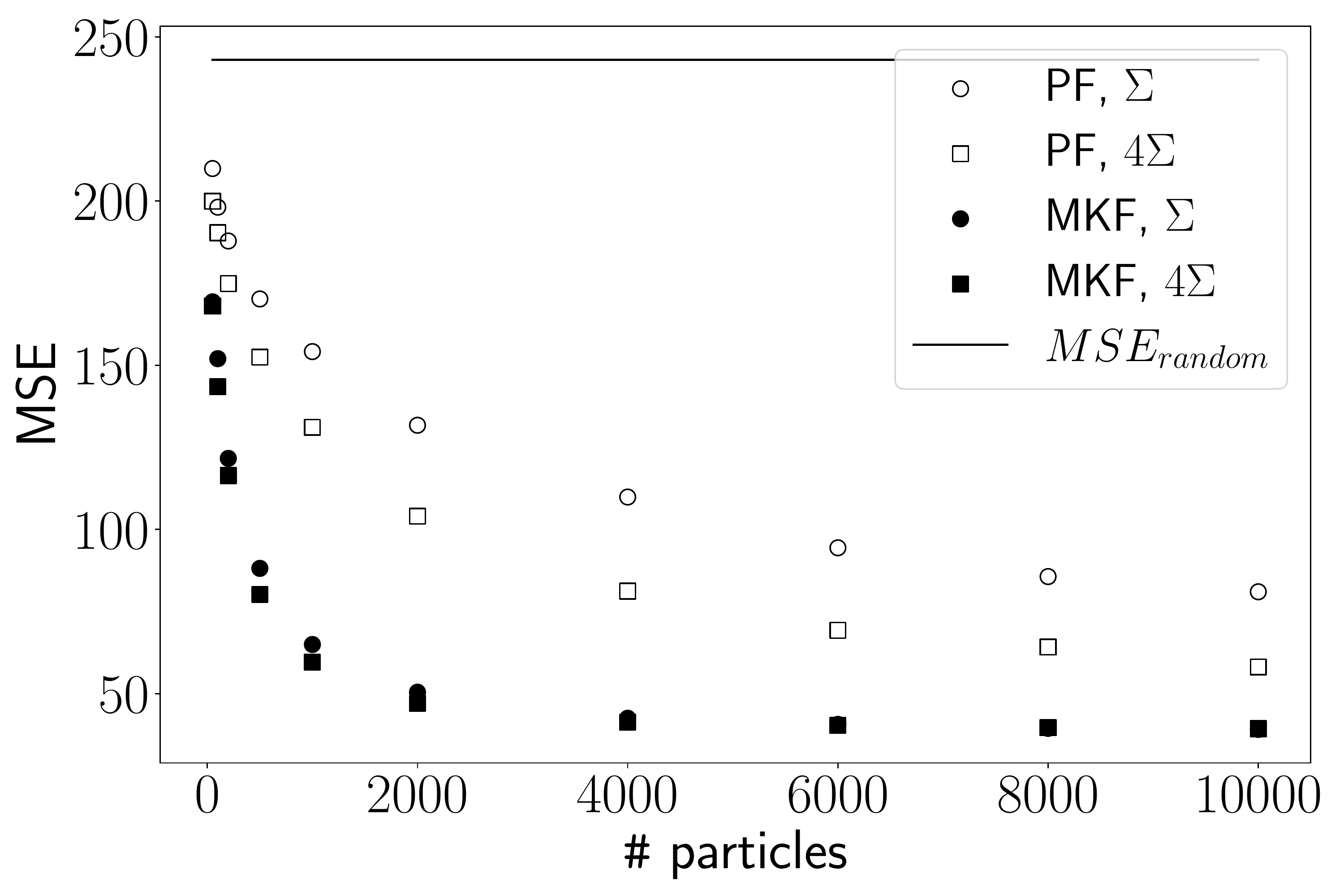}
    \caption{\textit{n-WORLD $27 \times 27$}}
    \end{subfigure}
\caption{Dependence of MSE on the number of particles in three non-symmetric environments. 
Multiparticle Kalman filter (MKF) demonstrates more accurate filtering of states and requires fewer particles for MSE convergence compared to the particle filter (PF). 
Our method is also less sensitive to the estimate of the motion noise than the particle filter.}
\label{fig::MSE1}
\end{figure}

The comparison results of the proposed filtering method with the particle filter in terms of the MSE~\eqref{eq::mse_def} are shown in Figures~\ref{fig::MSE} and~\ref{fig::MSE1} for symmetric and non-symmetric environments, respectively.
Both plots show that the proposed filtering method (MKF) requires fewer particles to achieve smaller values of MSE in the considered environments.
Also, one can observe that the filtering process in non-symmetric environments is more accurate and robust than in symmetric environments.
The smaller value of MSE indicates higher accuracy and the robustness is illustrated by the number of particles necessary for the convergence of MSE.
Also, these plots show that the proposed method is less sensitive to the estimate of motion noise than the particle filter.

\begin{figure}[!ht]
    \begin{subfigure}{0.3\textwidth}
    \includegraphics[width=\textwidth]
    {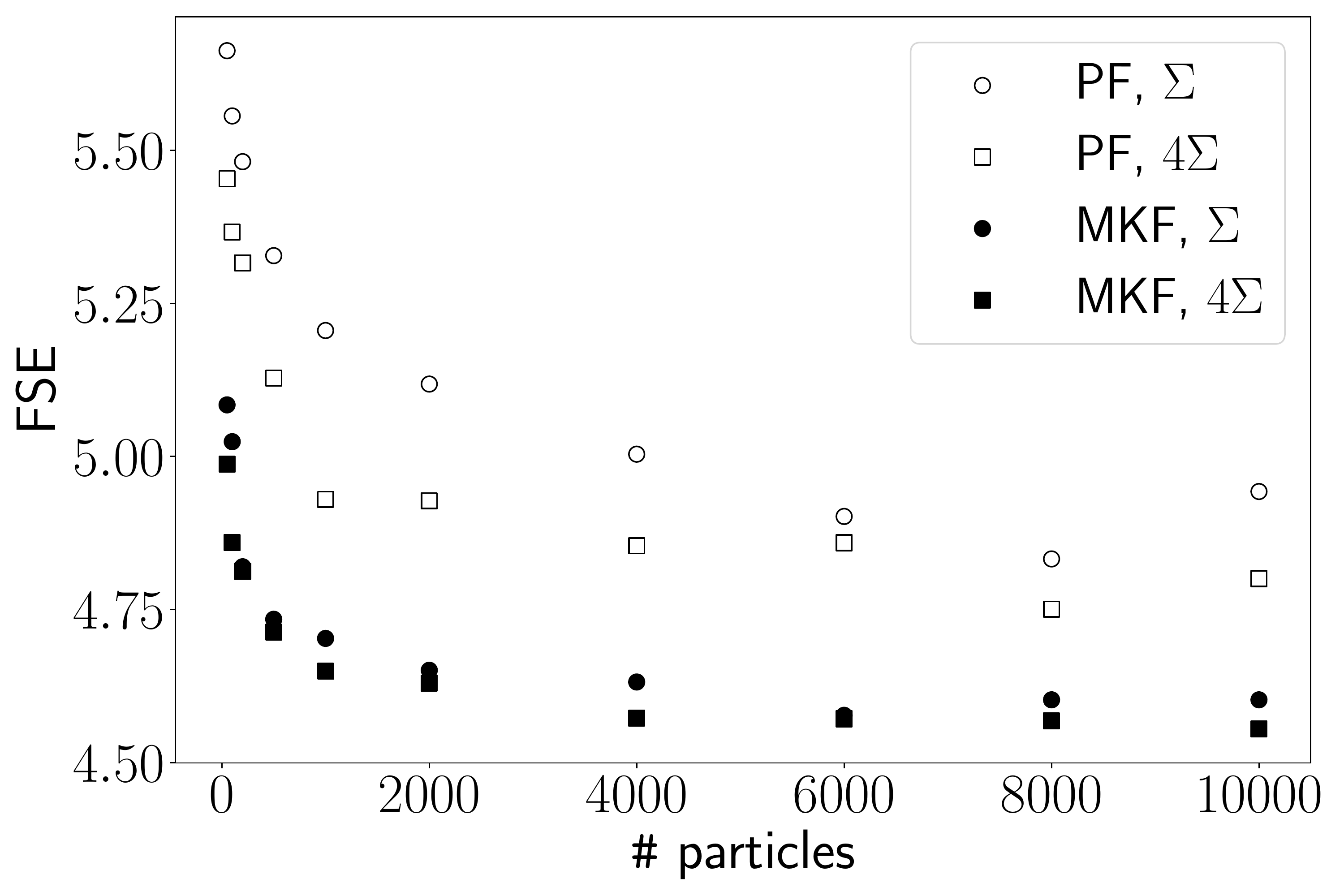}
    \caption{\textit{world $10 \times 10$}}
    \end{subfigure}
    ~
    \begin{subfigure}{0.3\textwidth}
    \includegraphics[width=\textwidth]
    {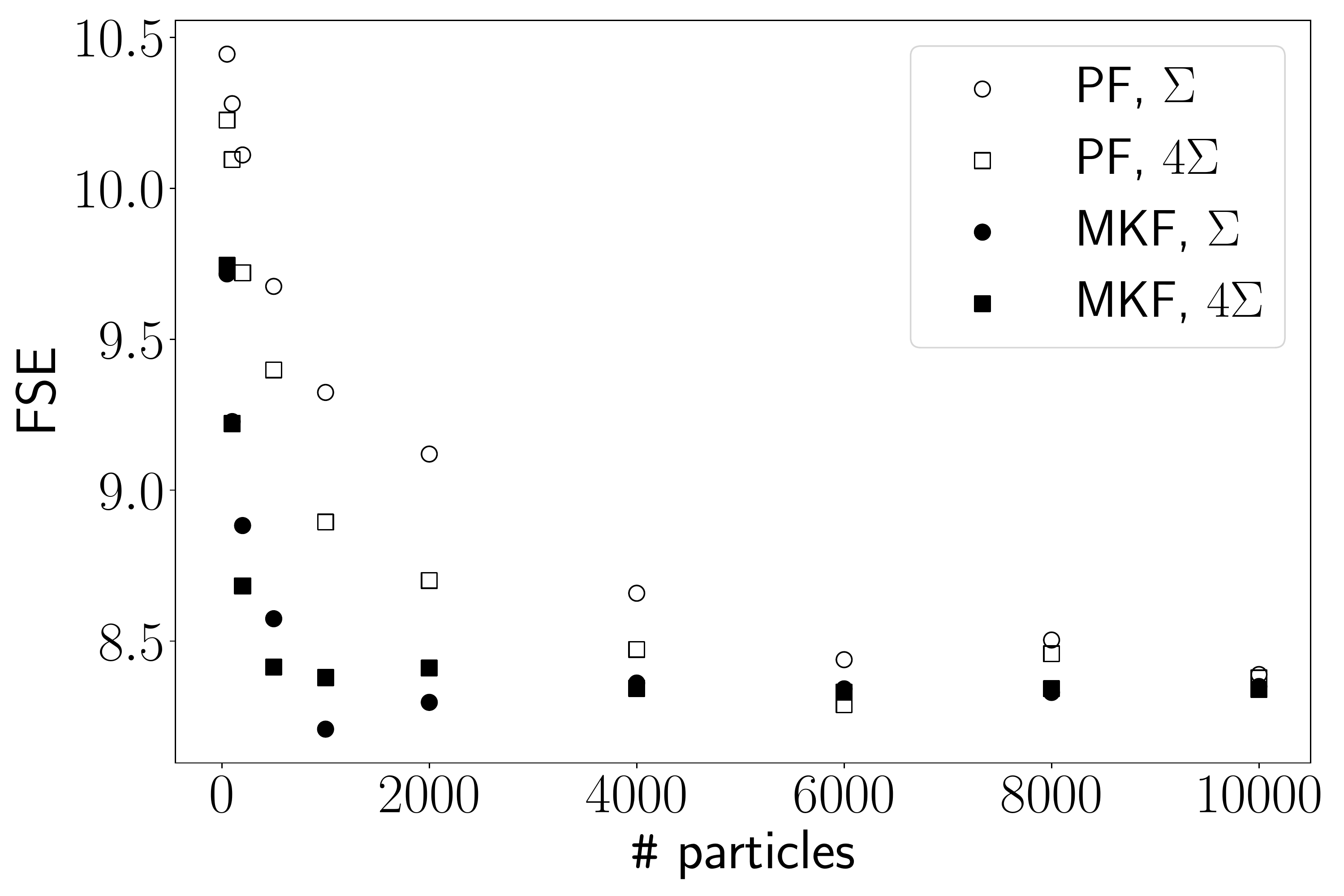}
    \caption{\textit{World $18 \times 18$}}
    \end{subfigure}
~
    \begin{subfigure}{0.3\textwidth}
    \centering
    \includegraphics[width=\textwidth]
    {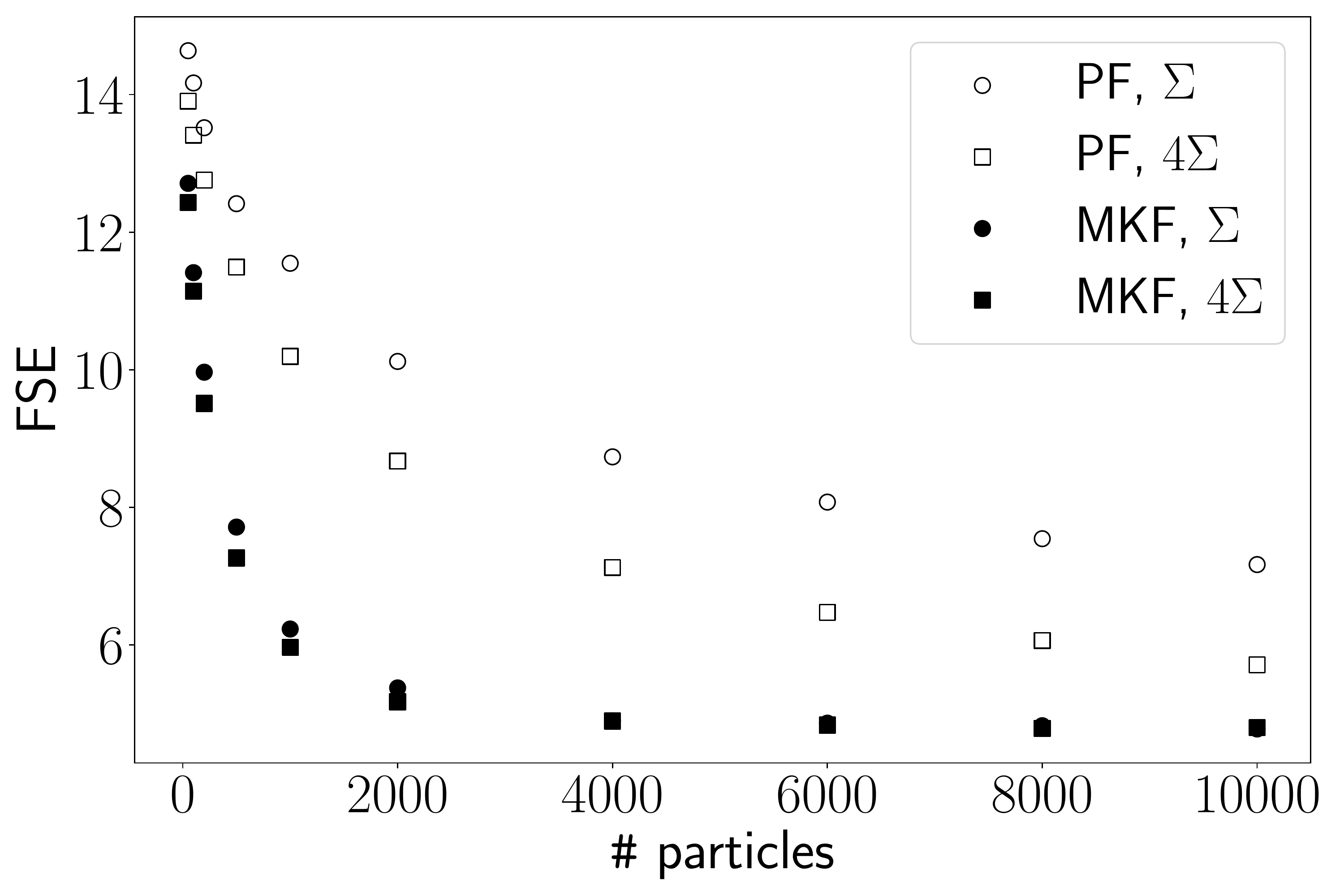}
    \caption{\textit{WORLD $27 \times 27$}}
    \end{subfigure}
\caption{Dependence of the final error loss function (FSE) on the number of particles used in the PF and MKF in the considered symmetric environments. MKF provides more accurate filtering of the states and requires fewer particles for convergence of FSE. Our filtering method is also less sensitive to the estimate of the motion noise than the particle filter.}
\label{fig::Fin}
\end{figure}

\begin{figure}[!ht]
    \begin{subfigure}{0.3\textwidth}
    \includegraphics[width=\textwidth]
    {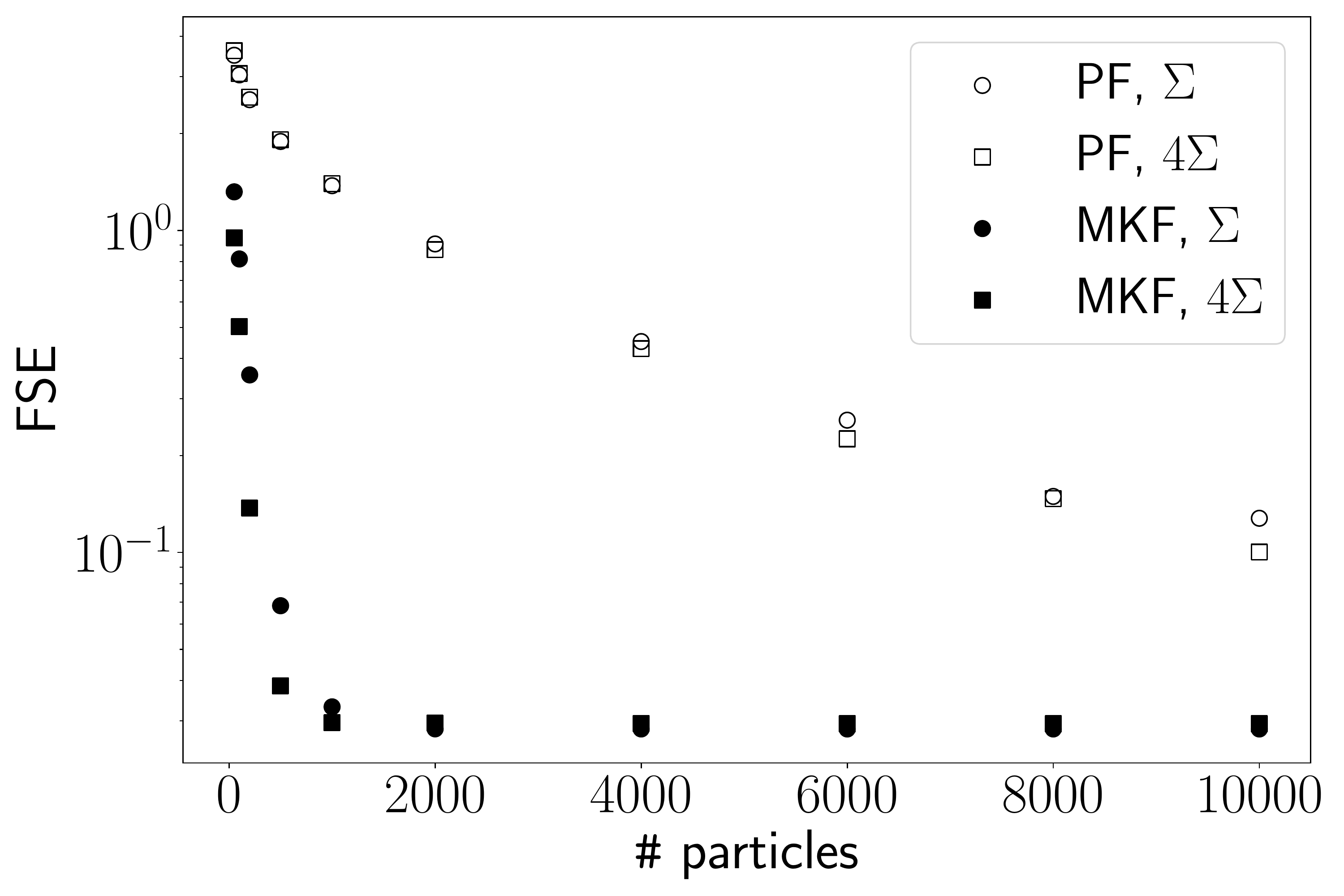}
    \caption{\textit{n-world $10 \times 10$}}
    \end{subfigure}
    ~
    \begin{subfigure}{0.3\textwidth}
    \includegraphics[width=\textwidth]
    {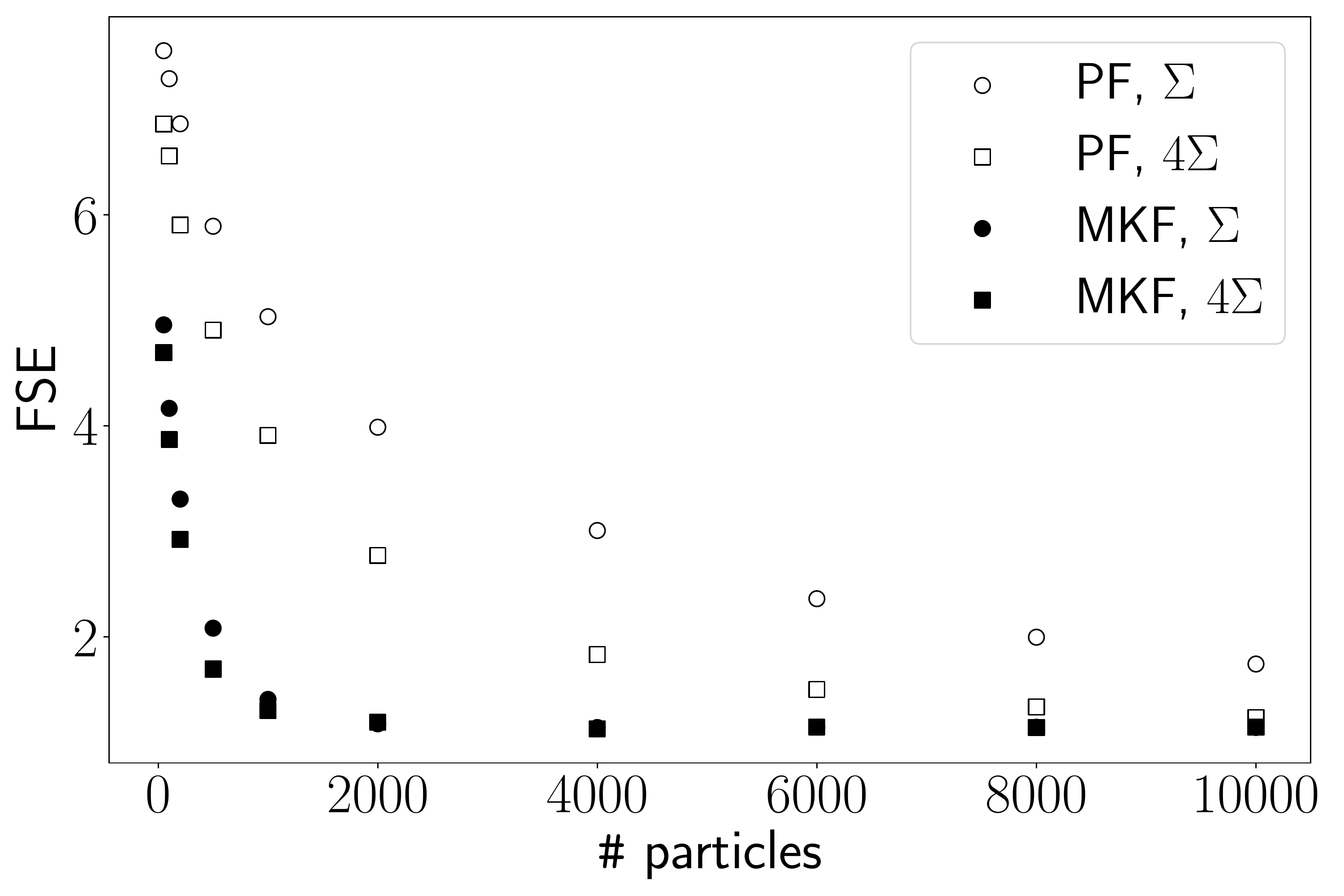}
    \caption{\textit{n-World $18 \times 18$}}
    \end{subfigure}
~
    \begin{subfigure}{0.31\textwidth}
    \centering
    \includegraphics[width=\textwidth]
    {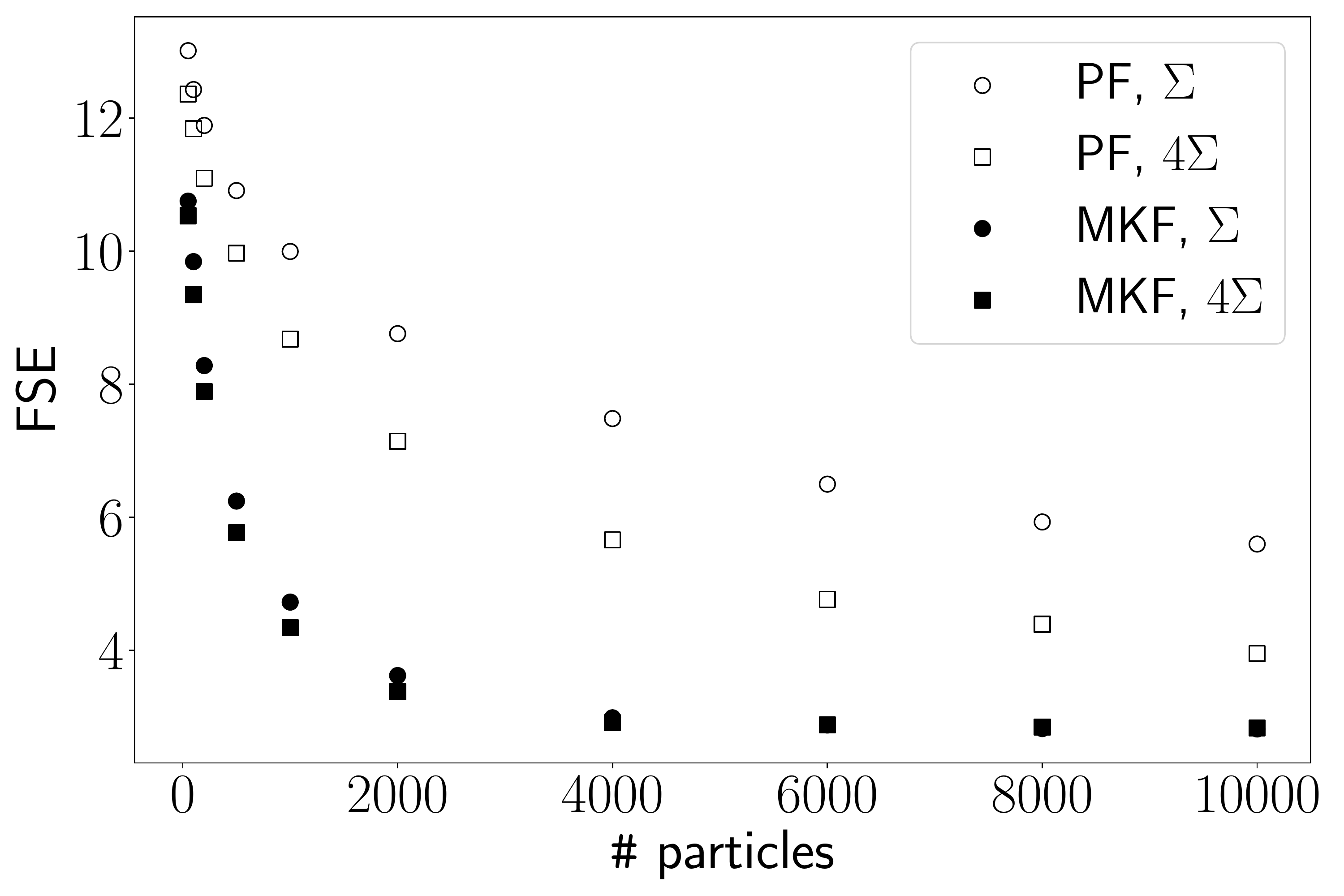}
    \caption{\textit{n-WORLD $27 \times 27$}}
    \end{subfigure}
\caption{Dependence of the final error loss function (FSE) on the number of particles used in the PF and MKF in the considered non-symmetric environments. 
MKF provides more accurate filtering of the states and requires fewer particles for the convergence of FSE. Our method is also less sensitive to the estimate of the motion noise than the particle filter.}
\label{fig::Fin1}
\end{figure}

Additional experiments are carried out to evaluate the considered filtering methods in terms of the final state error function~\eqref{eq::fse_def}.
The comparison results are shown in Figures~\ref{fig::Fin} and~\ref{fig::Fin1} for symmetric and non-symmetric environments, respectively.
The final states are computed after $100$ time steps in the considered environments.
These plots demonstrate the same trends that are observed in the analysis of MSE dependence on the number of particles presented in Figures~\ref{fig::MSE} and~\ref{fig::MSE1}.

\begin{figure}[!ht]
\begin{subfigure}{0.45\textwidth}
\includegraphics[width=\textwidth]
    {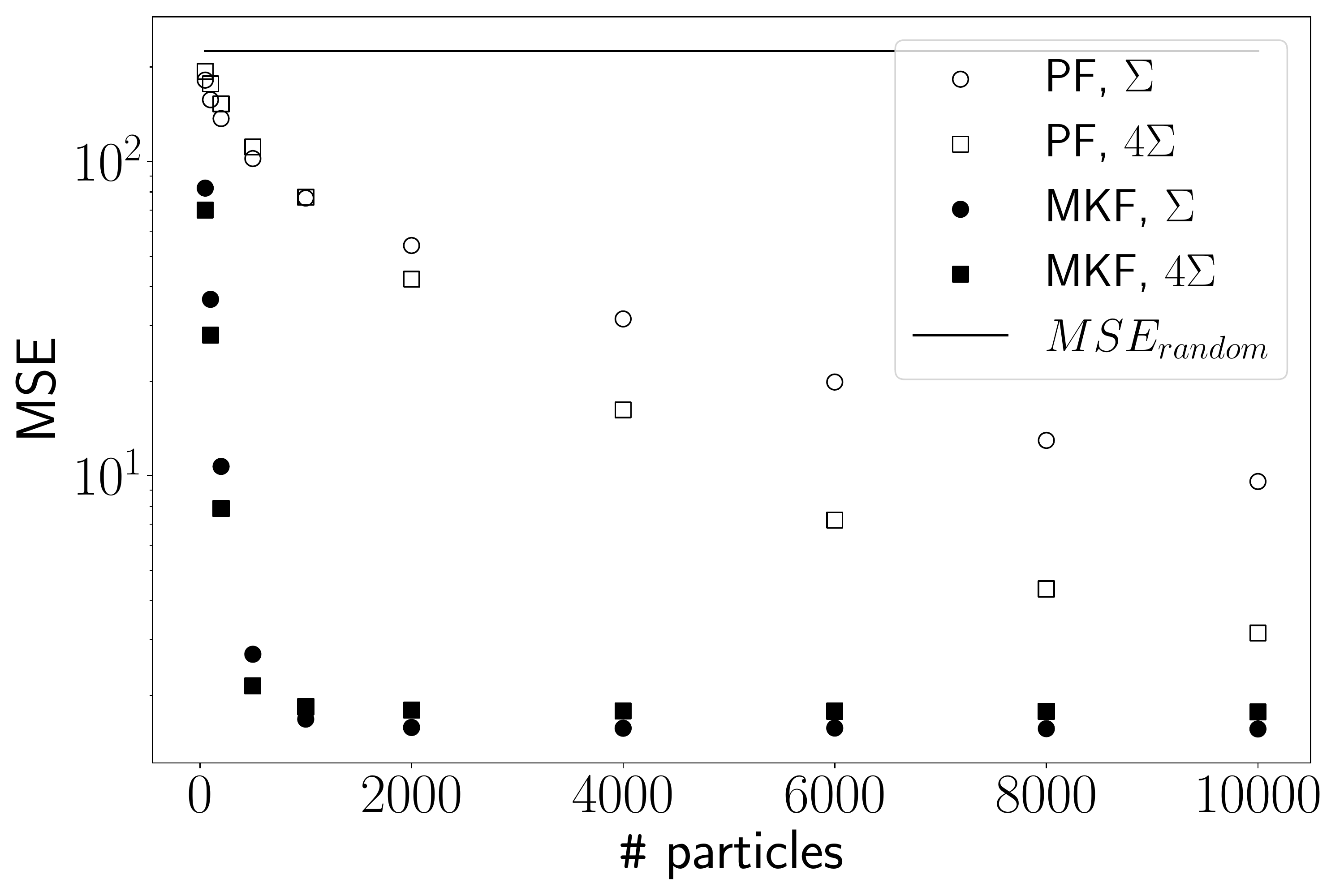}
\end{subfigure}
    ~
    \begin{subfigure}{0.45\textwidth}
\includegraphics[width=\textwidth]
    {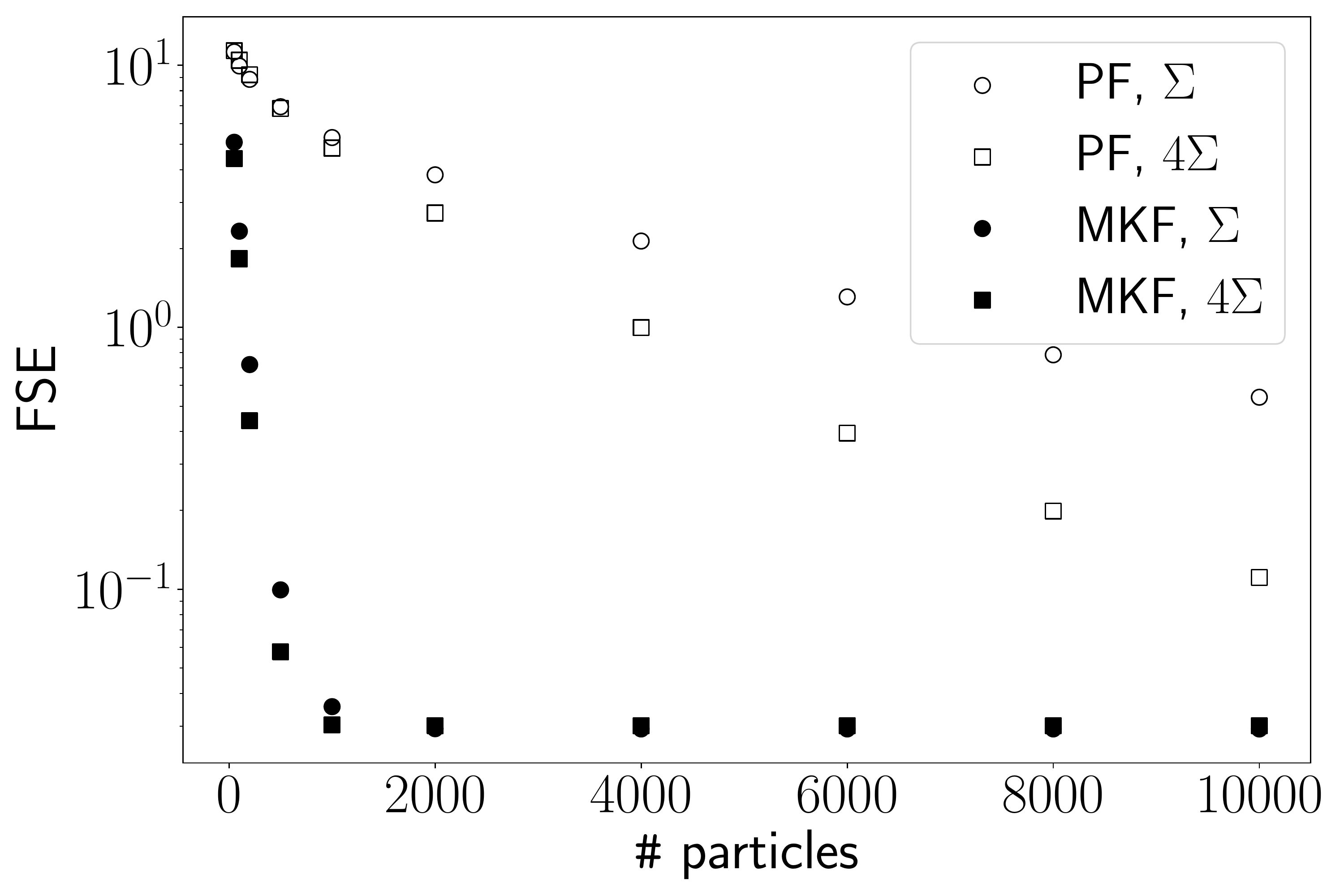}
\end{subfigure}
    \caption{Dependence of MSE (left) and FSE (right) values on the number of particles in the \emph{Labyrinth} environment. Our method (MKF) is also less sensitive to the estimate of the motion noise than the particle filter.}
    \label{fig::lab_mse_fse}
\end{figure}

Last but not least comparison of the particle filter and the multiparticle Kalman filter is performed in the \empty{Labyrinth} environment (see Figure~\ref{fig::labyrinth}).
Figure~\ref{fig::lab_mse_fse} shows that the proposed filtering method outperforms the particle filter in terms of both MSE and FSE quality criteria.
Also, we again observe the smaller number of particles required for the convergence of both loss functions.
The proposed method is more robust with respect to the motion noise level than the particle filter, which is aligned with previous results.

\paragraph{Variance analysis.}
To make the previous plots clear, we do not provide confidence intervals there.
To fill this gap in the reporting comparison results, we summarize the FSE values and the corresponding variance in Table~\ref{tab::variance_comparison}.
This table shows that the MKF provides a more accurate and less variable estimation of the final state for both symmetric and nonsymmetric environments.
This gain is observed uniformly with respect to the considered range of the number of particles.

\begin{table}[!h]
    \centering
    \caption{FSE values and variance comparison of particle filter (PF) and the proposed multiparticle Kalman filter (MKF). Standard deviation is given in braces near the corresponding mean FSE value. In these simulations, we use $\mM = 4\mM_0$, which is equal to $4\Sigma$ setting.}
    \begin{adjustbox}{width=\columnwidth,center}
    \begin{tabular}{ccccccccccc}
    \toprule
    Number of particles & \multicolumn{2}{c}{$N=100$} & \multicolumn{2}{c}{$N=500$} & \multicolumn{2}{c}{$N=1000$} & \multicolumn{2}{c}{$N=4000$} & \multicolumn{2}{c}{$N=10000$}\\
    \midrule
    Environment &  PF & MKF &  PF & MKF & PF & MKF & PF & MKF & PF & MKF \\
    \cmidrule(lr){1-1} \cmidrule(lr){2-3} \cmidrule(lr){4-5} \cmidrule(lr){6-7} \cmidrule(lr){8-9} \cmidrule(lr){10-11}
        \textit{world 10} & 5.37 (3.53) & 4.86 (3.76) & 5.13 (3.91) & 4.71 (2.78) & 4.93 (3.87) & 4.65 (2.44) & 4.85 (3.31) & 4.57 (2.05) & 4.8 (2.82) & 4.55 (1.98) \\
        \textit{n-world 10} & 3.07 (3.63) & 0.24 (1.26) & 1.91 (3.25) & 0.03 (0.07) & 1.40 (2.88) & 0.43 (1.67) & 0.04 (0.21) & 0.11 (0.70) &0.03 (0.02) & 0.03 (0.02) \\
        \textit{World 18} & 10.10 (6.49) & 9.22 (7.87) & 9.40 (7.72) & 8.41 (7.22) & 8.89 (8.20) & 8.38 (5.94) & 8.47 (7.79) & 8.34 (3.62) & 8.38 (6.64) & 8.34 (3.18) \\
        \textit{n-World 18} & 6.56 (6.94) & 3.87 (6.37) & 4.91 (6.89) & 1.70 (4.54) & 3.91 (6.46) & 1.30 (3.75) & 1.83 (4.86) & 1.13 (2.99) & 1.24 (3.77) & 1.15 (2.94) \\
        \textit{WORLD 27} & 13.41 (7.50) & 11.14 (7.62)  & 11.49 (7.48) & 7.26 (6.33) & 10.19 (7.21) & 5.96 (5.55) & 7.13 (6.27) & 4.89 (4.42) & 5.71 (5.47) & 4.80 (4.23) \\
        \textit{n-WORLD 27} & 11.84 (8.60) & 9.35 (8.41) & 9.97 (8.32) & 5.77 (6.86) & 8.68 (8.03) & 4.34 (5.89) & 5.66 (6.75) & 2.91 (4.39) & 3.95 (5.56) & 2.83 (4.25) \\
        \emph{Labyrinth} & 10.45 (10.11) & 1.83 (5.65) & 6.84 (9.53) & 0.06 (0.68) & 4.83 (8.49) & 0.03 (0.02) & 1.00 (4.23) & 0.03 (0.02) & 0.11 (1.17) & 0.03 (0.02) \\
        \bottomrule
    \end{tabular}
    \end{adjustbox}
    \label{tab::variance_comparison}
\end{table}



\paragraph{Runtime comparison.}
In the previous sections, we demonstrate the performance of the proposed filtering method in terms of the required number of particles for convergence of MSE and FSE and the smaller variance of these quantities compared to the particle filter.
Here, we provide the runtime comparison of the proposed filtering method and the particle filter.
In this experiment, we simulate $10000$  trajectories in the considered environments and provide the total runtime of such a simulation. 
Since the runtime of both compared methods significantly depends on the used number of particles, we consider $2000$ and $5000$ particles in the particle filter simulations and report the resulting FSE values.
Then, we tune the number of particles in the MKF such that the resulting FSE values are the same or slightly smaller than the corresponding FSE in the particle filter simulations.
The measured runtime, FSE, and the numbers of particles are shown in Table~\ref{tab::time_comparison}.
From this Table follows that the proposed multiparticle Kalman filter is typically 3-4 times faster than the particle filter.
This observation indicates that the gain from the reduction of the number of particles dominates the increasing per-iteration complexity of the proposed method.


\begin{table}[!h]
    \centering
    \caption{Comparison of the total runtime of particle filter (PF) and the proposed multiparticle Kalman filter (MKF) to simulate 10000 trajectories in the considered environments. The number of particles required for the MKF is set such that it achieves the same or slightly smaller FSE compared to values from PF simulations.}
    \begin{adjustbox}{width=0.8\columnwidth,center}
    \begin{tabular}{ccccccc}
    \toprule
     & \multicolumn{3}{c}{PF} & \multicolumn{3}{c}{MKF} \\
    \cmidrule(lr){2-4} \cmidrule(lr){5-7} 
    Environment & \# particles & FSE & Time, s & \# particles & FSE & Time, s\\
    \midrule
        \textit{world 10} & 2000 & 5.03 & 156 & 100 & 4.92 & 58 \\
        \textit{n-world 10} & 2000 & 0.87 & 151 & 100 & 0.80 & 46 \\
        \textit{World 18} & 2000 & 9.26 & 186 & 200 & 8.96 & 80 \\
        \textit{n-World 18} & 2000 & 4.87 & 168 & 100 & 4.61 & 50 \\
        \textit{WORLD 27} & 2000 & 11.0 & 204 & 200 & 10.62 & 86 \\
        \textit{n-WORLD 27} & 2000 & 9.58 & 211 & 150 & 9.45 & 70 \\
        \emph{Labyrinth} & 2000 & 4.8 & 139 & 100 & 3.02 & 48 \\
        \midrule
        \textit{world 10} & 5000 & 4.72 & 368 & 150 & 4.84 & 63 \\
        \textit{n-world 10} & 5000 & 0.30 & 334 & 250 & 0.25 & 90 \\
        \textit{World 18} & 5000 & 8.81 & 415 & 200 & 0.20 & 75 \\
        \textit{n-World 18} & 5000 & 3.49 & 412 & 300 & 3.20 & 111 \\
        \textit{WORLD 27} & 5000 & 9.50 & 489 & 400 & 9.00 & 153 \\
        \textit{n-WORLD 27} & 5000 & 7.94 & 473 & 400 & 7.54 & 151 \\
        \emph{Labyrinth} & 5000 & 2.69 & 323 & 150 & 1.86 & 61 \\
        \bottomrule
    \end{tabular}
    \end{adjustbox}
    \label{tab::time_comparison}
\end{table}

\section{Conclusion}



In the presented study, we consider the object localization problem with an unknown initial state in both symmetric and non-symmetric environments. 
We demonstrate that the standard particle filter algorithm performs poorly in highly symmetrical environments. 
We propose a novel multiparticle Kalman filter (MKF) based on the combination of the extended Kalman filter and particle filter. 
The MKF successfully addresses the problem of uncertainty in the object's initial state and outperforms the particle filter in all considered environments.
Our numerical experiments demonstrate that MKF requires fewer particles to achieve convergence in terms of both MSE and FSE quality criteria. 
Although every iteration of the proposed method is more costly compared to the particle filter, MKF converges faster since fewer particles are required to achieve the same error rates.
Also, we show that MKF is more robust to the level of measurement noise than the classical particle filter.

\section*{Acknowledgement}
This work is supported by the Ministry of Science and Higher Education of the Russian Federation (Grant 075-10-2021-068).

\bibliographystyle{unsrt}
\bibliography{lib}

\end{document}